\documentclass{article}

\usepackage{microtype}
\usepackage{graphicx}
\usepackage{subfigure}
\usepackage{booktabs} 

\usepackage{hyperref}

\usepackage[accepted]{icml2025}

\usepackage{amsmath}
\usepackage{amssymb}
\usepackage{mathtools}
\usepackage{amsthm}

\usepackage[capitalize,noabbrev]{cleveref}

\theoremstyle{plain}

\theoremstyle{definition}

\theoremstyle{remark}

\usepackage[textsize=tiny]{todonotes}

\usepackage{caption}
\usepackage{array}
\usepackage{tabularx}
\usepackage{makecell}
\usepackage{threeparttable} 

\usepackage{soul}
\usepackage{tcolorbox}
\usepackage{enumitem}
\usepackage{alltt}
\usepackage{listings}
\usepackage{fancyvrb}

\usepackage[utf8]{inputenc}
\usepackage{xcolor}
\tcbuselibrary{breakable,skins}
\usepackage{subcaption}

\newtcolorbox{promptbox}[2][]{
    colback=gray!5!white,
    colframe=gray!75!black,
    boxrule=0.5pt,
    fonttitle=\bfseries,
    title=#2,
    breakable,
    enhanced jigsaw,
    coltitle=black,
    colbacktitle=gray!30,
    left=1mm,
    right=1mm,
    top=1mm,
    bottom=1mm,
    #1
}

\makeatletter
\def\icml@appearance@address{Contact: } 
\makeatother

\icmltitlerunning{A2Eval: Agentic and Automated Evaluation for Embodied Brain}

\begin{document}

\twocolumn[
\icmltitle{A2Eval: Agentic and Automated Evaluation for Embodied Brain}



\icmlsetsymbol{equal}{*}

\begin{icmlauthorlist}
\icmlauthor{Shuai Zhang}{zju,westlake,robot}
\icmlauthor{Jiayu Hu}{robot}
\icmlauthor{Zijie Chen}{westlake}
\icmlauthor{Zeyuan Ding}{robot}
\icmlauthor{Yi Zhang}{robot}
\icmlauthor{Yingji Zhang}{manchester}
\icmlauthor{Ziyi Zhou}{robot,sustech}
\icmlauthor{Junwei Liao}{robot}
\icmlauthor{Shengjie Zhou}{xiamen}
\icmlauthor{Yong Dai}{robot}
\icmlauthor{Zhenzhong Lan}{westlake}
\icmlauthor{Xiaozhu Ju}{robot}
\end{icmlauthorlist}

\icmlaffiliation{zju}{Zhejiang University, China}
\icmlaffiliation{westlake}{Westlake University, China}
\icmlaffiliation{robot}{Beijing Innovation Center of Humanoid Robotics, China}
\icmlaffiliation{manchester}{University of Manchester, UK}
\icmlaffiliation{sustech}{Southern University of Science and Technology, China}
\icmlaffiliation{xiamen}{Xiamen University Malaysia, Malaysia}

\icmlcorrespondingauthor{Shuai Zhang}{zhangshuai@westlake.edu.cn}

\icmlkeywords{Machine Learning, ICML}

\vskip 0.3in
]



\printAffiliationsAndNotice{}  

\begin{abstract}
Current embodied VLM evaluation relies on static, expert-defined, manually annotated benchmarks that exhibit severe redundancy and coverage imbalance. This labor‑intensive paradigm drains computational and annotation resources, inflates costs, and distorts model rankings, ultimately stifling iterative development.
To address this, we propose Agentic Automatic Evaluation (A2Eval), the first agentic framework that automates benchmark curation and evaluation through two collaborative agents. The Data Agent autonomously induces capability dimensions and assembles a balanced, compact evaluation suite, while the Eval Agent synthesizes and validates executable evaluation pipelines, enabling fully autonomous, high-fidelity assessment. 
Evaluated across 10 benchmarks and 13 models, A2Eval compresses evaluation suites by 85\%, reduces overall computational costs by 77\%, and delivers a 4.6$\times$ speedup while preserving evaluation quality. Crucially, A2Eval corrects systematic ranking biases, improves human alignment to Spearman's $\rho=0.85$, and maintains high ranking fidelity (Kendall's $\tau=0.81$), establishing a new standard for high-fidelity, low-cost embodied assessment. 
Our code and data will be public soon.
\end{abstract}

\section{Introduction}

\begin{figure}[t!]
\centering

\begin{minipage}{\linewidth}
\centering
\includegraphics[width=\linewidth]{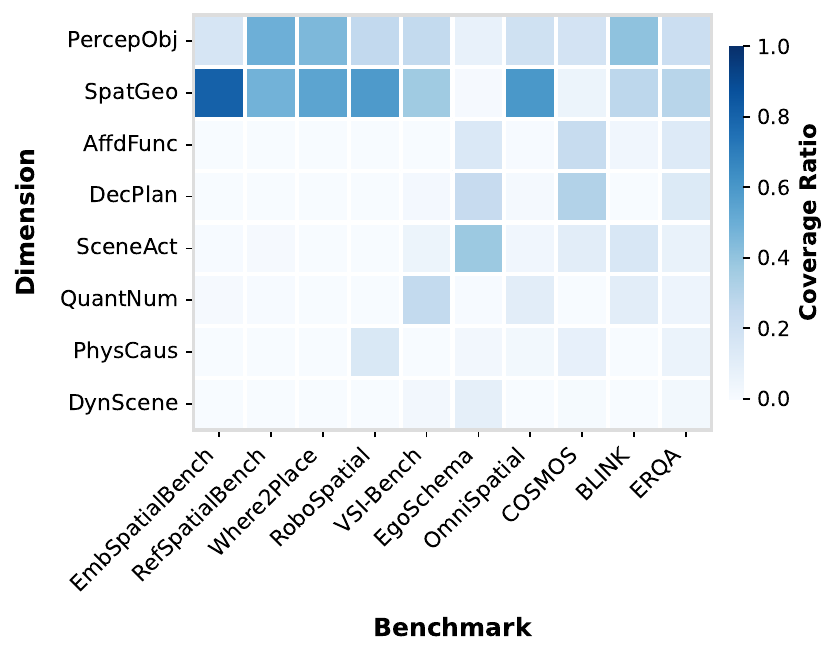}
\\ \small (a) Capacity coverage in embodied VLM benchmarks.
\end{minipage}

\vspace{1.0em}

\begin{minipage}{\linewidth}
\centering
\includegraphics[width=\linewidth]{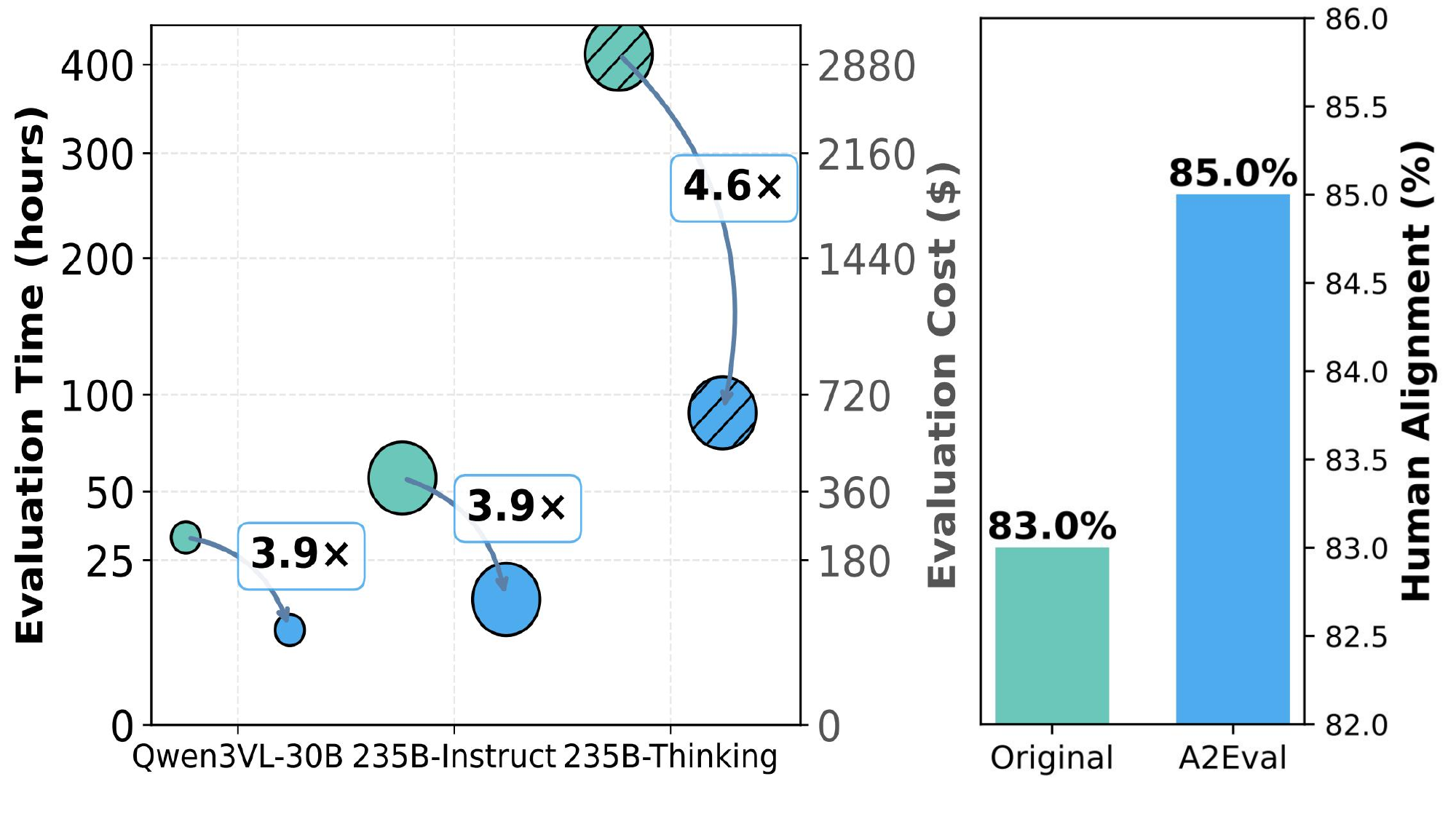}
\\ \small (b) Evaluation Cost and Human Alignment.
\end{minipage}

\caption{\textbf{(a)}: Capacity distributions across existing expert-defined and manually annotated embodied VLM benchmarks reveal redundant yet sparse coverage, leading to ranking distortion and excessive evaluation costs. \textbf{(b)}: A2Eval replaces this expert-defined, annotation-heavy paradigm with an automated suite that maintains capability coverage while compressing benchmarks, achieving 4.6$\times$ speedup and improved human alignment. (Details in Tables~\ref{tab:rank} and \ref{tab:model_efficiency}.)} 

\label{fig:main_teaser}
\end{figure}

Evaluating modern embodied vision-language models (VLM) has become the ``compass'' for navigating the rapid advancements in Embodied AI~\cite{liu2025embodied,cong2025overview}. A comprehensive and precise evaluation mechanism is not only a benchmark for academic breakthroughs but also a cornerstone for industrial decision-making~\cite{yifan2025embodied,zhang2025pelican}. 
However, the cost of this ``compass'' is becoming prohibitive: evaluating a single model across fragmented benchmarks now consumes over \textit{3200 GPU hours} (see Table~\ref{tab:model_efficiency}), creating a significant financial and computational burden that stifles iterative research.

\begin{figure*}[t]
  \centering
  \includegraphics[width=0.95\textwidth]{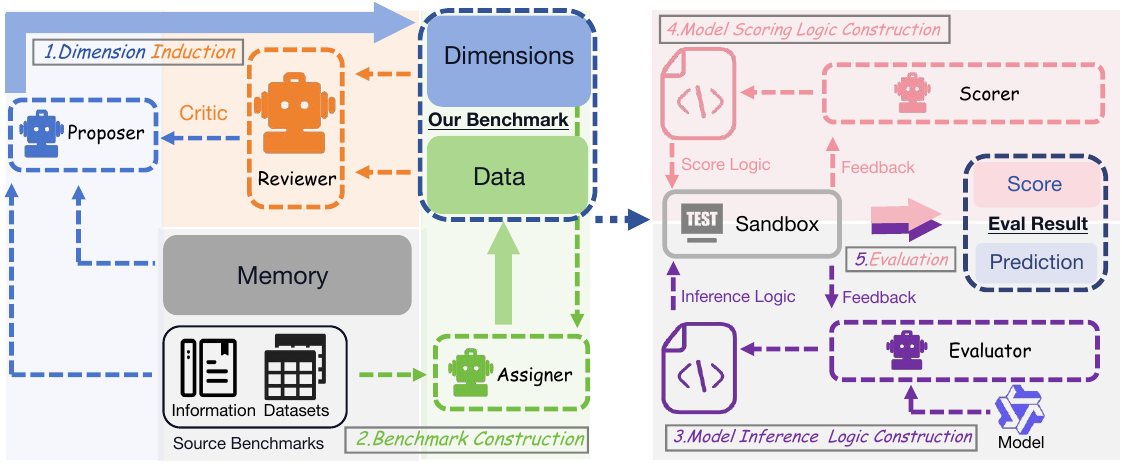}
  \caption{Overview of A2Eval. The \textbf{left} half shows the \textbf{Data Agent} with three roles, Proposer, Reviewer, and Assigner, which constructs compact, balanced benchmarks. The \textbf{right} half shows the \textbf{Eval Agent} with two roles, Evaluator and Scorer, which produces model predictions and scores. Each role is represented by a distinct color.}
  \label{fig:pipeline}
\end{figure*}

This crisis stems from a \textit{Broken Evaluation Ecosystem} rooted in the labor-intensive \textbf{``expert definition + manual annotation''} paradigm. As shown in Figure~\ref{fig:main_teaser}(a) and \ref{fig:bmk_heatmap}, current benchmarks exhibit up to \textit{92\%} sample similarity and severe capacity distribution skew. Critical reasoning capacity like \textit{PhysCausal} constitute less examples, while easy-to-annotate tasks dominate. This creates a ``lose-lose'' dilemma characterized by three systemic pathologies:
\textbf{(1) Coverage Imbalance \& Redundancy:} Massive sample duplication and skewed task distribution lead to repetitive, unbalanced assessments that fail to comprehensively evaluate model capabilities.
\textbf{(2) Ranking Distortion:} Models overfit to overrepresented tasks appear top-ranked despite critical weaknesses, distorting leaderboards and misleading research directions.
\textbf{(3) Prohibitive Evaluation Cost:} The combination of evaluating massive redundant samples and manually constructing inference and scoring logic for each benchmark creates substantial computational and human effort overhead, making evaluation cycles slow and costly, stifling iterative development.

To resolve these tensions, we propose \textit{Agentic and Automatic Evaluation (A2Eval)}, a paradigm shift that treats benchmark curation as an optimization problem: \textit{maximize capability coverage and diversity while minimizing redundancy and human effort}.
A2Eval replaces the traditional paradigm with a fully autonomous process via two collaborative agents:

\textbf{Data Agent} addresses \textit{Coverage Imbalance}, \textit{Redundancy}, and \textit{Ranking Distortion} by autonomously inferring capability dimensions—eliminating reliance on subjective expert definitions. By automatically inducing capability dimensions and performing diversity-aware sampling, it constructs a balanced, compact evaluation suite that substantially reduces benchmark size and redundant computation while maintaining comprehensive capacity coverage, thereby correcting systematic ranking biases and significantly lowering evaluation cost.

\textbf{Eval Agent} resolves the \textit{Manual Pipeline Bottleneck} by automatically synthesizing and validating executable evaluation pipelines. Through iterative sandbox refinement of both inference and scoring logic, it enables fully autonomous, high-fidelity evaluation without manual intervention.

Extensive experiments across 10 benchmarks and 13 models demonstrate the effectiveness of our approach. Our main contributions are summarised as follows:

\textbf{(1) First Agentic Evaluation Framework:} We present the first fully automated evaluation framework that decouples assessment from static expert labels, replacing manual processes with a scalable, autonomous agentic system that addresses both benchmark construction and execution challenges.

\textbf{(2) Balanced and Compact Benchmark Construction:} Our Data Agent corrects \textit{Coverage Imbalance} and \textit{Ranking Distortion}, filters \textit{85\%} redundant samples, boosts human preference alignment to \textit{0.85} (Spearman's $\rho$), and reduces overall cost by \textit{77\%} with \textit{4.6$\times$} evaluation throughput acceleration.

\textbf{(3) Autonomous Executable Evaluation Pipeline:} Our Eval Agent automatically constructs validated inference and scoring logic, achieving \textit{96.9\%} fidelity while enabling fully autonomous evaluation without manual intervention.

\section{Related Work}

\subsection{Embodied Reasoning Benchmarks}

\begin{table*}[htbp]
\centering
\small
\setlength{\tabcolsep}{9pt}
\renewcommand{\arraystretch}{1}
\begin{tabular}{p{0.26\textwidth} p{0.60\textwidth}}
\toprule
\textbf{Dimension} & \textbf{Definition} \\
\midrule

Perception \& Object Grounding (\texttt{PercepObj}) &
Localizing, identifying, and tracking objects, such as selecting coordinates or masks, detecting bounding boxes, and choosing grasp points. \\[6pt]

Scene \& Action Understanding (\texttt{SceneAct}) &
Interpreting scenes, actions, states, and poses, such as recognizing environments, human or agent activities, object states, and body or pose cues. \\[6pt]

Spatial \& Geometric Reasoning (\texttt{SpatGeo}) &
Understanding spatial relations, frames of reference, and topological structures, including position, orientation, containment, and adjacency. \\[6pt]

Quantitative \& Numerical Reasoning (\texttt{QuantNum}) &
Determining or comparing numerical properties, including counting items, estimating continuous quantities, and judging relative magnitudes. \\[6pt]

Affordance \& Function Reasoning (\texttt{AffdFunc}) &
Assessing what actions objects afford, such as grasping, opening or pouring, tool use, and feasibility of manipulation under constraints. \\[6pt]

Physical \& Causal Reasoning (\texttt{PhysCaus}) &
Reasoning about physical feasibility and stability, such as judging fit, support, balance, and predicting collisions or other physical interactions. \\[6pt]

Decision \& Task Planning (\texttt{DecPlan}) &
Planning and decision-making across step,s including choosing next actions, applying conditional logic and decomposing tasks into sub-goals. \\[6pt]

Dynamic Scene Reasoning (\texttt{DynScene}) &
Understanding how objects and the whole scene change, interact, and stay consistent over time, such as event order, state and spatial relations. \\

\bottomrule
\end{tabular}
\caption{Data agent induced capability dimensions (left: full name and abbreviation) with detailed definitions (right).}
\label{tab:dimension_definitions}
\end{table*}

Recent benchmarks evaluate a wide range of embodied reasoning capabilities. Tasks such as Where2Place~\cite{yuan2025robopoint} and VSI-Bench~\cite{yang2025thinking} target placement and spatial understanding, while RefSpatial~\cite{zhou2025roborefer}, RoboSpatial~\cite{song2025robospatial}, EmbSpatial~\cite{du2024embspatial}, and OmniSpatial~\cite{jia2025omnispatial} extend spatial reasoning to grounding and manipulation. ERQA~\cite{team2025gemini} and COSMOS~\cite{azzolini2025cosmos} focus on physical and causal reasoning, whereas BLINK~\cite{fu2024blink} and EgoSchema~\cite{mangalam2023egoschema} assess multi-modal perception and egocentric understanding. Recent vision–language models (e.g., Qwen3-VL~\cite{Qwen3-VL}, InternVL3.5~\cite{wang2025internvl3}) and embodied-specialized systems (e.g., Pelican-VL~\cite{zhang2025pelican, zhang2025bridging}, RoboBrain~\cite{team2025robobrain}) increasingly incorporate such benchmarks, alongside system-level frameworks like EmbodiedBench~\cite{yang2025embodiedbench} and EAI~\cite{li2024embodied}. However, the rapid proliferation of benchmarks introduces evaluation challenges, including substantial redundancy~\cite{zhang2025redundancy} and high computational cost, particularly for long video-based benchmarks~\cite{mangalam2023egoschema, yang2025thinking}.

\subsection{Agentic Evaluation and Benchmarks}
Parallel efforts have explored automated benchmark construction. Code2Bench~\cite{zhang2025dynamic} and OKBench~\cite{li2025okbench} provide domain-specific pipelines for generating new benchmarks with contamination control and reproducibility protocols. However, these methods focus on creating fresh evaluation sets rather than consolidating existing heterogeneous benchmarks or addressing redundancy and computational costs.
Our work differs from prior work by introducing (1) automatic capability induction with balanced sampling across domains, (2) redundancy-aware benchmark consolidation, and (3) a unified end-to-end data and evaluation pipeline.

\section{Method}

We propose Agentic Automatic Evaluation (A2Eval) to address the challenges of benchmark redundancy, fragmented coverage, and high evaluation costs in embodied domains. As shown in Figure \ref{fig:pipeline}, A2Eval is a two-agent framework that automates the complete evaluation lifecycle from benchmark construction to model assessment. It consists of a Data Agent (Section \ref{dataagent}), which discovers unified capability dimensions through specialized agent collaboration (\textbf{Proposer, Reviewer, and Assigner}) and constructs a compact, balanced benchmark through the Assigner's dimension assignment and diversity-aware sampling, and an Eval Agent (Section \ref{evalagent}), which generates executable evaluation pipelines through iterative code synthesis. For detailed algorithmic procedures and prompt designs, refer to Appendix~\ref{app:pipeline}.

\subsection{Data Agent}
\label{dataagent}
The Data Agent constructs a capability-aware, low-redundancy, and balanced benchmark for a target domain given existing benchmark information pools $\mathcal{B} = \{B_i\}_{i=1}^{N_B}$ and their example sets $\mathcal{E}=\bigcup_{i=1}^{N_B} E_i$, where $N_B$ is the number of benchmarks. The core innovation lies in treating benchmark construction as a two-stage agentic workflow: (i) \textbf{Dimension Induction}, where multiple specialized agents collaboratively discover a unified capability taxonomy, and (ii) \textbf{Benchmark Construction}, where the Assigner performs dimension assignment and diversity-aware sampling to produce a balanced, non-redundant test set that preserves comprehensive semantic coverage.

\paragraph{Dimension Induction.}
Dimension induction operates through iterative collaboration among three specialized agents with distinct roles and reasoning capabilities: a \textbf{Proposer} $\mathcal{A}_p$ that generates candidate capability dimensions, a \textbf{Reviewer} $\mathcal{A}_r$ that critiques proposals from coverage, independence and balance perspectives, and an \textbf{Assigner} $\mathcal{A}_a$ that reports dimension balance through example assignment and diversity-aware sampling. These agents communicate through a shared memory $M$ that records historical proposals and critiques, enabling collective reasoning toward an optimal dimension set.

At iteration $i$, the Proposer generates a candidate dimension set $D_{i+1}$ by reasoning over benchmark information $\mathcal{B}$, the Reviewer's critique $C_i$, and historical memory $M$:
\begin{equation}
D_{i+1} = \mathcal{A}_p(\mathcal{B}, C_i, M).
\end{equation}
The \textbf{Proposer} refines dimensions based on specific feedback from $C_i$ while incorporating experience from $M$.

The \textbf{Reviewer} evaluates $D_{i+1}$ along three criteria: (a) conceptual redundancy among dimensions, (b) coverage comprehensiveness relative to domain requirements, and (c) dimension balance. The critique $C_{i+1}$ is produced by:
\begin{equation}
C_{i+1} = \mathcal{A}_r(D_{i+1}, S^{(r)}).
\end{equation}
where $S^{(r)}$ contains balance statistics from the Assigner. The memory is updated as $M \leftarrow M \cup \{(D_{i+1}, C_{i+1})\}$ after each iteration.

The Proposer and Reviewer iterate until dimensions stabilize ($D_{i+1} = D_i$), producing candidate $D^{(r+1)}$. To validate balance, $D^{(r+1)}$ is tested through actual example assignment and sampling by the Assigner, yielding balance statistics $S^{(r+1)}$. The Reviewer incorporates this empirical evidence into its next critique, triggering dimension refinement (e.g., merging under-represented dimensions). This agent-driven validation loop continues until producing a stable dimension set $D$.

\paragraph{Benchmark Construction.}
Given the validated dimension set $D$, the \textbf{Assigner} constructs the benchmark through a two-stage process: dimension assignment followed by diversity-aware sampling. This transforms the dimension taxonomy into an actual test set that eliminates redundancy while preserving comprehensive capability coverage.

\textbf{Dimension Assignment.} The Assigner employs an ensemble of $N_v$ voter agents $\{\mathcal{A}_a^{1}, \dots, \mathcal{A}_a^{N_v}\}$ that independently reason about capability requirements. Each voter $\mathcal{A}_a^k$ analyzes example $e_j \in \mathcal{E}$ and produces dimension assignment $D_{jk}$ based on its interpretation. The final assignment $D_j$ emerges from majority consensus:
\begin{equation}
D_j = \text{MajorityVote}\big(\{D_{jk}\}_{k=1}^{N_v}\big).
\end{equation}
Per-dimension example pools are aggregated as 
\begin{equation}
E_{d_k} = \{e_j \in \mathcal{E} : d_k \in D_j\}
\end{equation}
for each $d_k \in D$.

\textbf{Diversity-Aware Sampling.} Following dimension assignment, the Assigner performs diversity-aware sampling to eliminate intra-dimension redundancy while preserving semantic diversity through clustering-based selection. For each dimension $d_k$, examples $e \in E_{d_k}$ are embedded as $x = \text{Encoder}(e)$ into a semantic space.
If $|E_{d_k}| < K$, all examples are retained; otherwise, we select the example closest to each of the $K$ cluster centroids $\mu_j$:
\begin{equation}
E_{d_k}^{\mathrm{s}} = \big\{\arg\min_{x\in C_j}\|x-\mu_j\|\big\}_{j=1}^{K}.
\label{eq:diversity_sampling}
\end{equation}
This centroid-based selection ensures that retained examples $E_{d_k}^{\mathrm{s}}$ sparsely yet strategically cover the semantic manifold (visualized in Figure~\ref{fig:PercepObj_SpatGeo}), removing redundancy without sacrificing representation completeness. By applying uniform $K$ across dimensions, cross-dimension balance is achieved while each dimension maintains maximal internal diversity. The final benchmark is constructed as:
\begin{equation}
\mathcal{E}^{\text{s}} = \bigcup_{k=1}^{N_d} E_{d_k}^{\mathrm{s}}.
\end{equation}
with balance statistics
\begin{equation}
S = \{|E_{d_k}^{\text{s}}| : d_k \in D\}.
\end{equation}

\newcolumntype{L}{>{\raggedright\arraybackslash}X}
\begin{table}[t]
\centering
\footnotesize
\setlength{\tabcolsep}{4pt}
\renewcommand{\arraystretch}{1}
\begin{tabularx}{\columnwidth}{LLL}
\toprule
\textbf{Dimension} & \textbf{Source Set} & \textbf{Retained Set} \\
\midrule
PercepObj   & 6,218 (25.4\%)  & 500 (13.2\%) \\
SceneAct    & 1,864 (7.6\%)   & 500 (13.2\%) \\
SpatGeo     & 10,565 (43.1\%) & 500 (13.2\%) \\
QuantNum    & 3,185 (13.0\%)  & 500 (13.2\%) \\
AffdFunc    & 773 (3.2\%)     & 500 (13.2\%) \\
PhysCausal  & 366 (1.5\%)     & 366 (9.7\%)  \\
DecPlan     & 1,104 (4.5\%)   & 500 (13.2\%) \\
DynScene    & 425 (1.7\%)     & 425 (11.2\%) \\
Other       & 19 (0.0\%)      & 0 (0.0\%)    \\
\midrule
\textbf{Total} & \textbf{24,519} & \textbf{3,781} \\
\bottomrule
\end{tabularx}
\caption{Number of examples per capability dimension before and after fixed-$K$ cluster sampling.
``Source Set'' denotes examples labeled by the data agent;
``Retained Set'' denotes examples retained after diversity-aware sampling.
Percentages are shown in parentheses.}
\label{tab:dimension_stat}
\end{table}

\subsection{Eval Agent}
\label{evalagent}
The Eval Agent executes automated evaluation on the benchmark produced by the Data Agent. It consists of two specialized roles: (i) an \textbf{Evaluator} that constructs model inference logic, and (ii) a \textbf{Scorer} that constructs performance assessment logic. These roles collaborate with a sandbox executor to iteratively generate and validate a complete evaluation pipeline.

\paragraph{Model Inference Logic Construction.}
The \textbf{Evaluator} $\mathcal{A}_{e}$ synthesizes inference logic that integrates the target model $M$ with benchmark test examples $\mathcal{T}$ sampled from $\mathcal{E}^{\text{s}}$. At iteration $i$, the Evaluator generates executable inference code $F_i^e$ that implements: loading test inputs, invoking $M$, and producing model predictions.

The sandbox executor $\mathcal{X}$ runs $F_i^e$ on model $M$ with test cases $\mathcal{T}$. If execution succeeds, the validated logic $F^e = F_i^e$ is finalized. Otherwise, $\mathcal{X}$ returns diagnostic feedback $E_i$ to the Evaluator to update inference logic:
\begin{equation}
F_{i+1}^e = \mathcal{A}_e(M, \mathcal{T}, F_{i}^e, E_i).
\end{equation}
This synthesis-validation loop iterates until $F^e$ executes reliably or reaches a predefined iteration budget. The resulting artifact $F^e$ constitutes the validated inference logic for systematic model evaluation.

\paragraph{Model Scoring Logic Construction.}
Given the validated inference logic $F^e$ and the model predictions $\mathcal{P}_{\mathcal{T}} = F^e(M, \mathcal{T})$ it produces on test cases $\mathcal{T}$, the \textbf{Scorer} $\mathcal{A}_{s}$ synthesizes evaluation logic to assess model performance. At iteration $j$, the Scorer generates executable scoring code $F_j^s$ conditioned on the test cases $\mathcal{T}$ and model outputs $\mathcal{P}_{\mathcal{T}}$:

The sandbox executor $\mathcal{X}$ runs $F_j^s$ on the test outputs $\mathcal{P}_{\mathcal{T}}$. If execution succeeds and produces valid metrics, the logic $F^s = F_j^s$ is finalized. Otherwise, $\mathcal{X}$ returns diagnostic feedback $E_j$ to the Scorer, triggering iterative refinement:
\begin{equation}
F_{j+1}^s = \mathcal{A}_s(\mathcal{T}, \mathcal{P}_{\mathcal{T}}, F_{j}^s, E_j).
\end{equation}
The synthesis-validation loop continues until $F^s$ executes reliably or reaches the iteration budget. The resulting artifact $F^s$ constitutes reproducible scoring logic for comprehensive model assessment.

\paragraph{Benchmark Evaluation Execution.}
Once both inference logic $F^e$ and scoring logic $F^s$ are validated, the complete evaluation pipeline is executed on the constructed benchmark $\mathcal{E}^{\text{s}}$. The Evaluator applies $F^e$ to generate model predictions for all test cases:
\begin{equation}
\mathcal{P} = F^e(M, \mathcal{E}^{\text{s}}),
\end{equation}
and the Scorer applies $F^s$ to compute final evaluation metrics:
\begin{equation}
\text{Results} = F^s(\mathcal{E}^{\text{s}}, \mathcal{P}).
\end{equation}
This produces comprehensive assessment results for the target model $M$ across all capability dimensions in $D$.

\begin{table}[t]
\centering
\footnotesize
\setlength{\tabcolsep}{4pt}
\renewcommand{\arraystretch}{1}

\begin{tabularx}{\columnwidth}{l>{\centering\arraybackslash}X
                                  >{\centering\arraybackslash}X
                                  >{\centering\arraybackslash}X}
\toprule
\textbf{Model} &
$\widetilde{\mathcal{B}}$ &
$\mathcal{B}$ &
$\mathcal{H}$ \\
\midrule
Qwen3-VL-235B-A22B-Thinking    & 65.97 & 61.26 & 87.9 \\
Internvl-3.5-241B-A28B         & 65.68 & 56.76 & 63.6 \\
GPT-5-20250807-Mini            & 65.52 & 55.16 & 79.4 \\
Qwen3-VL-30B-A3B-Thinking      & 62.58 & 54.14 & 71.2 \\
Internvl-3.5-38B               & 62.37 & 55.01 & 64.5 \\
Qwen3-VL-235B-A22B-Instruct    & 61.88 & 55.39 & 77.4 \\
Internvl-3.5-30B-A3B           & 61.79 & 51.38 & 55.3 \\
Qwen3-VL-30B-A3B-Instruct      & 61.47 & 48.36 & 62.5 \\
Internvl-3.5-8B                & 55.98 & 48.84 & 61.5 \\
Qwen2.5-VL-72B-Instruct        & 54.76 & 51.07 & 61.3 \\
Qwen2.5-VL-32B-Instruct        & 50.77 & 47.24 & 58.5 \\
Qwen2.5-VL-7B-Instruct         & 48.22 & 44.26 & 55.5 \\
Qwen2.5-VL-3B-Instruct         & 39.39 & 37.62 & 50.4 \\
\midrule
\multicolumn{4}{l}{\textit{Ranking correlation with Human ($\mathcal{H}$)}} \\
\multicolumn{1}{l}{\quad Spearman's $\rho$} & \textbf{0.85} & 0.83 \\
\multicolumn{1}{l}{\quad Kendall's $\tau$}  & \textbf{0.72} & 0.64 \\
\midrule
\multicolumn{4}{l}{\textit{Ranking correlation between Our ($\widetilde{\mathcal{B}}$) and Source ($\mathcal{B}$})} \\
\multicolumn{1}{l}{\quad Spearman's $\rho$} & \multicolumn{3}{c}{0.94} \\
\multicolumn{1}{l}{\quad Kendall's $\tau$}  & \multicolumn{3}{c}{0.81} \\
\bottomrule
\end{tabularx}
\caption{Model scores on our constructed benchmark $\widetilde{\mathcal{B}}$, the original benchmark $\mathcal{B}$, and human evaluations $\mathcal{H}$. The bottom section reports rank correlations with human evaluations and between the two benchmarks.}
\label{tab:rank}
\end{table}

\section{Experiments}
\label{sec:exp}
We evaluate A2Eval in the embodied vision--language reasoning domain, where models must jointly perform perception, spatial reasoning, physical understanding, and multi-step decision making over visual inputs. We validate that our Data Agent produces semantically meaningful capability dimensions and balanced benchmarks, aligning with human judgment rankings. Moreover, we highlight that our approach significantly improves evaluation efficiency, while the Eval Agent achieves high accuracy in automated inference and scoring. Together, these results demonstrate that A2Eval enables fully automated, efficient, and trustworthy capability-aware evaluation.

\subsection{Experimental Setup}
\paragraph{Benchmarks.}
We focus on the embodied vision--language reasoning domain and adopt ten established benchmarks, compromising 
COSMOS \cite{azzolini2025cosmos}, ERQA \cite{team2025gemini}, Where2Place \cite{yuan2025robopoint}, VSI-Bench \cite{yang2025thinking},
OmniSpatial \cite{jia2025omnispatial}, EgoSchema \cite{mangalam2023egoschema}, BLINK \cite{fu2024blink}, 
RefSpatial \cite{zhou2025roborefer}, RoboSpatial \cite{song2025robospatial}, and EmbSpatial \cite{du2024embspatial}.
These datasets both inform benchmark construction via the Data Agent and serve as reference corpora for analyzing redundancy, coverage, and evaluation efficiency. Detailed information on these benchmarks and their redundancy examples is provided in Appendix~\ref{app:benchmark}.

\paragraph{Models.}
We evaluate a diverse set of vision language models with different scales and architectures, including
several Qwen2.5-VL models \cite{bai2025qwen2},
multiple InternVL-3.5 variants \cite{wang2025internvl3},
the latest Qwen3-VL series \cite{Qwen3-VL},
as well as the GPT-5 Mini as a closed-source model.
All evaluations are performed with a batch size of 1 using the lmms-eval framework \cite{zhang2024lmmsevalrealitycheckevaluation}.

\paragraph{Evaluation Settings.}
By default, large-scale models (e.g., Qwen3-VL-235B-A22B-Thinking) are deployed with the vLLM backend, while smaller models (e.g., Qwen2.5-VL-7B-Instruct) utilize the PyTorch backend with 8-way data parallelism via Accelerate.
We apply greedy decoding for all models to ensure deterministic outputs.
Video inputs are uniformly sampled to 32 frames at 2 FPS, and the maximum generation length is set to 4096 tokens.
More details are provided in Appendix~\ref{app:eval_setting}.

\paragraph{A2Eval Setup.}
For the agentic workflow, we employ GPT-4o as the Assigner, while the Proposer, Reviewer, and Eval Agents are implemented using Gemini 3 Pro.
In the Assigner's diversity-aware sampling for the Data Agent, examples are encoded by concatenating visual and textual embeddings. Textual content is embedded using a sentence representation model\footnote{\url{https://huggingface.co/sentence-transformers/all-MiniLM-L6-v2}}, while visual representations are extracted using CLIP \cite{radford2021learning}. For video inputs, 32 uniformly sampled frames are processed, with the final visual embedding obtained by averaging their frame-level CLIP features. The number of clusters $K$ is set to 500 across all dimensions to strike a balance between semantic coverage and evaluation efficiency.

\subsection{Overall Results}
\paragraph{Unified Dimensions.} Our Data Agent automatically induces eight capability dimensions for embodied visual--language reasoning.
Table~\ref{tab:dimension_definitions} lists each induced dimension with a concise abbreviation and the full definition produced by the agent.
\paragraph{Balanced Benchmark.}
Table~\ref{tab:dimension_stat} reports the number of examples assigned to each dimension and retained after Assigner's diversity-aware sampling. The ``Other'' category is omitted due to negligible size. The raw pool contains 24,519 examples, while the sampled benchmark contains 3,781 examples.

\begin{figure}[ht]
\centering

\begin{minipage}{0.9\linewidth}
\centering
\includegraphics[width=\linewidth]{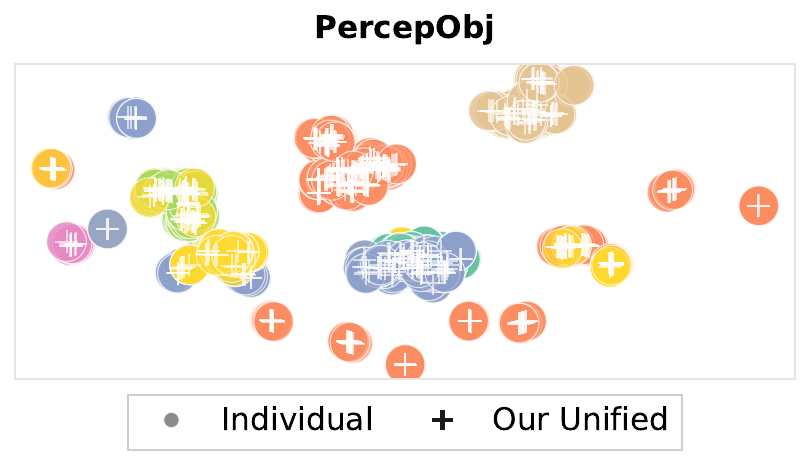}
\captionof{subfigure}{Perception object grounding module.}
\label{fig:percep}
\end{minipage}

\vspace{0.6em}

\begin{minipage}{0.9\linewidth}
\centering
\includegraphics[width=\linewidth]{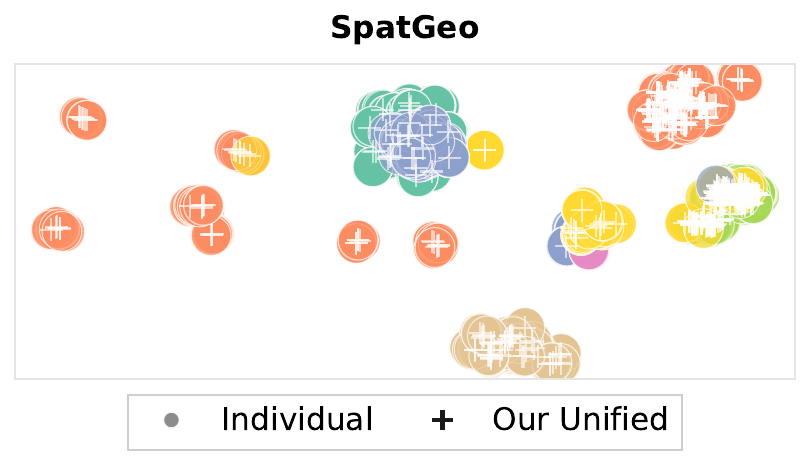}
\captionof{subfigure}{Spatial geometric reasoning module.}
\label{fig:spat}
\end{minipage}

\caption{UMAP visualization of sample embeddings, which demonstrates that our unified subset maintains comprehensive semantic coverage despite high sparsity.}
\label{fig:PercepObj_SpatGeo}
\end{figure}

\paragraph{Coverage and Diversity.}
Figure~\ref{fig:main_teaser}(a) visualizes how original benchmarks distribute across dimensions. Benchmarks emphasize distinct capabilities, with substantial overlap across dimensions. This reveals that aggregating existing benchmarks provides broad but redundant coverage, motivating capability-aware consolidation rather than simple union.
Figure~\ref{fig:PercepObj_SpatGeo} shows UMAP~\cite{mcinnes2018umap} projections of sample embeddings for two representative dimensions: PercepObj and SpatGeo (others are listed in Figure~\ref{fig:reasoning_modules}). Our selected subset (marked with “+” symbols) sparsely yet strategically covers the full semantic manifold, demonstrating that \ul{our Data Agent preserves comprehensive capability coverage while removing intra-dimension redundancy.}

\subsection{Quality of Induced Capability Dimensions}
We examine whether the capability dimensions induced by the \textbf{Data Agent} are
(i) conceptually well-formed as a representation of embodied video--language reasoning, and
(ii) reliably assigned to individual samples in practice.
\paragraph{Dimension Construction Quality.}
We assess the induced capability space along two axes:
\emph{comprehensiveness} (whether dimensions cover the core competencies required for embodied brain domain) and \emph{independence} (whether they capture distinct reasoning signals).

\textbf{Comprehensiveness.} The eight dimensions systematically decompose the cognitive requirements for embodied agents, spanning the processing hierarchy from sensory input to action generation. At the \emph{perceptual level}, PercepObj facilitates entity grounding, while SceneAct decodes semantic context and agent behaviors. Moving to the \emph{reasoning level}, SpatGeo resolves geometric relations and PhysCaus ensures adherence to physical laws and stability; meanwhile, AffdFunc assesses manipulation feasibility and QuantNum computes numerical properties. Finally, at the \emph{executive level}, DecPlan orchestrates action sequencing and goal decomposition, supported by DynScene which monitors temporal evolution and state transitions. \ul{This hierarchical organization aligns with the functional architecture of embodied brain systems, where vision-language models must bridge perception, reasoning, and planning for real-world interaction.}
Assignment statistics in Table~\ref{tab:dimension_stat} validate this coverage: among 24,519 examples, only 19 were classified as ``Other'', \ul{indicating the taxonomy captures the embodied domain without auxiliary categories.}

\textbf{Independence.} \ul{Dimensions are distinguished by reasoning mechanisms rather than input modalities or task surfaces.} For instance, SpatGeo (topological relations) and PercepObj (entity localization) both process visual input but invoke distinct cognitive operations; one requires relational abstraction while the other demands perceptual grounding. Similarly, PhysCaus (passive dynamics) and AffdFunc (action-oriented affordances) both involve physical understanding but differ in whether the agent reasons about natural laws versus intervention feasibility.


\subsection{Benchmark Validity and Rationality}
Beyond coverage and redundancy reduction, we examine that our constructed benchmark, (i) yields dimension assignments aligned with human judgment, (ii) preserves ranking fidelity with the full multi-benchmark setting, and (iii) produces a more rational, capability-aware ranking that better reflects human-perceived embodied competence.

\paragraph{Human--Agent Agreement on Dimension Assignment.}
To assess the alignment between automatic dimension assignments and human understanding, we conduct an agreement study with 5 domain experts (Appendix~\ref{app:human_agent_agreement}). Annotators review model-proposed labels in a blind setting, with final decisions determined by majority voting. We observe strong consistency between automatic and human-validated annotations, achieving an average Cohen’s $\kappa$ of 0.78 and an inter-annotator agreement of 0.80. \ul{These results confirm that our benchmark yields dimension assignments that are well aligned with human judgment.}

\paragraph{Ranking Fidelity under Compression.}
We compare model rankings from our constructed benchmark $\widetilde{\mathcal{B}}$ with those from all original benchmarks $\mathcal{B}$.
Table~\ref{tab:rank} shows that our method achieves strong alignment ($\rho=0.94$, $\tau=0.81$), indicating that \ul{redundancy removal substantially reduces evaluation volume (from 24,519 to 3,781 examples) while preserving the global comparative structure.}

\paragraph{Capability-Aware Ranking via Human Alignment.}
Although overall rankings remain stable, rebalancing corrects the relative ordering of models favored by dataset bias or oversampled capabilities. To verify that these shifts reflect improved evaluation rather than random variation, we construct a \textit{human-centric ranking} based on 8 realistic embodied scenarios (Appendix~\ref{app:human_eval_protocol}), including household, grocery store, laboratory, warehouse, and clinical environment. For each scenario, annotators watch real-world videos and assess models’ step-by-step embodied action plans under anonymized conditions. Each output is scored along eight capability-aligned criteria that mirror the induced dimensions, using a 1--5 Likert scale. The resulting human-perceived ranking serves as an independent reference to evaluate the alignment between capability-aware rebalancing and human judgment.

Table~\ref{tab:rank} reports high alignment between our benchmark
and human rankings ($\rho=0.85$, $\tau=0.72$), surpassing the original benchmark's
human correlation ($\rho=0.83$, $\tau=0.64$). Models improving in rank demonstrate stronger cross-dimension reasoning and consistent human preference.
This indicates \ul{our benchmark produces a more rational, capability-aware ordering
reflecting genuine embodied competence rather than dataset artefacts.}

\paragraph{Case Study of Qwen3-VL-30B-A3B-Instruct vs. Qwen2.5-VL-72B-Instruct.}
As shown in Table~\ref{tab:rank} and Table~\ref{tab:model_scores}, the original $\mathcal{B}$ ranks Qwen2.5-VL-72B-Instruct higher (51.07 vs. 48.36), but $\widetilde{\mathcal{B}}$ reverses this (61.47 vs. 54.76), aligning with human preference (62.5 vs. 61.3).
The reversal stems from correcting \textit{imbalanced dimension coverage}.
In $\mathcal{B}$, critical reasoning capabilities like PhysCaus (1.5\%) and
DecPlan (4.5\%) are under-tested, while SpatGeo dominates (43.1\%).
Our rebalancing (PhysCaus: 9.7\%, DecPlan: 13.2\%, SpatGeo: 13.2\%) reveals
large gaps: Qwen3-VL-30B-A3B-Instruct outperforms by 21.5 points on PhysCaus
(67.10 vs. 45.60) and 11.1 points on DecPlan (66.30 vs. 55.16), weaknesses
barely impacting aggregate scores in $\mathcal{B}$ due to low weight.
\ul{The rebalanced ranking now reflects consistent cross-dimension strength, matching human judgment.}

\newcolumntype{L}{>{\raggedright\arraybackslash}X}
\begin{table}[t]
\centering
\footnotesize
\setlength{\tabcolsep}{4pt}
\renewcommand{\arraystretch}{1}
\begin{tabularx}{\columnwidth}{p{3.3cm}>{\centering\arraybackslash}X>{\centering\arraybackslash}X}
\toprule
\textbf{Strategy} & \textbf{Agn. Source} & \textbf{Agn. Human} \\
\midrule
Source-proportional & \textbf{0.98} & 0.81 \\
Dimension-aware only & 0.92 & 0.82 \\
\midrule
\textbf{Ours (Dim. + Div.)} & 0.94 & \textbf{0.85} \\
\bottomrule
\end{tabularx}
\caption{Ablation study on Data Agent's sampling strategies. 
Ours (Dim. + Div.) denotes our agentic dimension-aware and diversity-aware sampling approach.
Spearman's $\rho$ measures ranking correlation with original benchmark scores (Agn. Source) and human evaluations (Agn. Human).}
\label{tab:ablation_sampling}
\end{table}

\paragraph{Ablation Study of Sampling Strategy.}
Table~\ref{tab:ablation_sampling} shows how each component of our agentic sampling of original benchmarks contributes to evaluation quality. \textbf{Source-proportional} sampling preserves the original benchmark distribution, achieving the highest $\mathcal{B}$ correlation ($\rho=0.98$) but the lowest human alignment ($\rho=0.81$), indicating that simply copying the original structure maintains its biases. 
\textbf{Dimension-aware only} adds agentic dimension construction with assignment via multi-agent voting while using uniform sampling within each dimension, improving human alignment to $\rho=0.82$ by balancing examples across dimensions. 
\textbf{Ours} (dimension-aware + diversity-aware) further incorporates diversity-aware sampling within each dimension, achieving the best human alignment ($\rho=0.85$) while maintaining reasonable $\mathcal{B}$ consistency ($\rho=0.94$). Both components are essential: \ul{agentic dimension awareness through multi-agent voting balances examples across dimensions, while diversity awareness through clustering-based sampling removes intra-dimension redundancy without sacrificing semantic coverage.}

\subsection{Computational Efficiency}
\ul{Our benchmark substantially reduces evaluation cost while preserving comprehensive capability assessment.} Table~\ref{tab:model_efficiency} reports wall-clock evaluation time for 7B–241B models. The benchmark yields consistent 3.4$\times$ to 4.6$\times$ speedups across all models. For Qwen3-VL-235B-A22B-Thinking, evaluation time decreases from 412.9 to 89.4 hours (4.6$\times$), while InternVL-3.5-241B-A28B reduces from 80.0 to 18.9 hours (4.2$\times$). This efficiency gain is critical for frontier-scale models, where full multi-benchmark evaluation becomes prohibitively expensive. Even smaller models like Qwen2.5-VL-7B-Instruct  achieve a 3.4$\times$ speedup (2.4 to 0.7 hours), demonstrating \ul{consistent benefits across scales.}

\newcolumntype{L}{>{\raggedright\arraybackslash}X}
\begin{table}[t]
\centering
\footnotesize
\setlength{\tabcolsep}{3pt}
\renewcommand{\arraystretch}{1}

\begin{tabularx}{\columnwidth}{Lccc}
\toprule
\textbf{Model} &
\textbf{Source} &
\textbf{Ours} &
\textbf{S.D.} \\
\midrule
Qwen3-VL-235B-A22B-Thinking        & 412.9 & 89.4 & 4.6$\times$ \\
Qwen3-VL-235B-A22B-Instruct        & 56.0  & 14.5  & 3.9$\times$ \\
InternVL-3.5-241B-A28B             & 80.0  & 18.9  & 4.2$\times$ \\
Qwen3-VL-30B-Instruct              & 32.3  & 8.3  & 3.9$\times$ \\
Qwen2.5-VL-72B-Instruct            & 11.2  & 3.0   & 3.7$\times$ \\
InternVL-3.5-8B                    & 3.8   & 0.9   & 4.2$\times$ \\
Qwen2.5-VL-7B-Instruct             & 2.4   & 0.7   & 3.4$\times$ \\
\bottomrule
\end{tabularx}

\caption{Evaluation time (in hours) comparison using 8 GPUs. 
S.D. denotes the speedup of our method over the original benchmark. Our method achieves consistent 3.4$\times$ to 4.6$\times$ speedups across all models.}
\label{tab:model_efficiency}
\end{table}

\subsection{Eval Agent Fidelity}
\label{subsec:eval_agent_fidelity}
Beyond efficiency, we assess the correctness and faithfulness of the Eval Agent by comparing its scores to reference implementations.

\paragraph{Evaluation Protocol.}
We measure overall fidelity by computing the relative difference between agent-generated and reference evaluation scores: $\text{Fidelity} = 1 - \frac{|\text{Score}_{\text{agent}} - \text{Score}_{\text{ref}}|}{|\text{Score}_{\text{ref}}|}$, where $\text{Score}_{\text{agent}}$ represents the final score produced by the complete agent-generated evaluation pipeline (including both inference logic $F^f$ and scoring logic $F^s$), and $\text{Score}_{\text{ref}}$ represents the score from the reference implementation. We evaluate three models (Qwen2.5-VL-7B-Instruct, Qwen3-VL-8B-Instruct, InternVL-3.5-8B) across all eight capability dimensions.
\begin{table}[ht!]
\centering
\footnotesize
\setlength{\tabcolsep}{3pt}
\renewcommand{\arraystretch}{1}
\begin{tabular}{lcccc}
\toprule
\textbf{Dimension} &
\textbf{Q2.5-7B} &
\textbf{Q3-8B} &
\textbf{IVL3.5-8B} &
\textbf{Avg} \\
\midrule
PercepObj  & 92.0  & 93.7 & 94.1  & 93.3 \\
SceneAct   & 97.2  & 98.5 & 98.6  & 98.1 \\
SpatGeo    & 95.7  & 94.4 & 91.0  & 93.7 \\
QuantNum   & 90.5  & 97.4 & 93.8  & 93.9 \\
AffdFunc   & 99.2  & 96.5 & 100.0 & 98.6 \\
PhysCaus   & 98.4  & 96.4 & 98.1  & 97.6 \\
DecPlan    & 96.4  & 95.5 & 94.5  & 95.5 \\
DynScene   & 94.1  & 96.6 & 99.2  & 96.6 \\
\midrule
\textbf{Final Score} & 97.2 & 96.2 & 97.4 & \textbf{96.9} \\
\bottomrule
\end{tabular}
\caption{Overall fidelity of the Eval Agent across all capability dimensions. Values represent the fidelity between agent-generated evaluation scores and reference scores. The agent achieves 96.9\% average fidelity, demonstrating our reliable end-to-end evaluation.}
\label{tab:total_fidelity}
\end{table}
\paragraph{Fidelity Results.}
Table~\ref{tab:total_fidelity} presents the overall fidelity evaluation across all capability dimensions and models. The Eval Agent achieves 96.9\% average fidelity, demonstrating \ul{highly reliable generation of evaluation pipelines.} This end-to-end evaluation validates that \ul{the complete pipeline, from inference to scoring, produces results that closely match reference implementations.} Detailed breakdowns of inference and scoring fidelity components are provided in Appendix~\ref{appendix:fidelity}.

\section{Conclusion}
In this work, we propose Agentic Automatic Evaluation (A2Eval), the first agentic framework for automated benchmark curation using two collaborative agents. The Data Agent autonomously induces capability dimensions to construct balanced evaluation suites, while the Eval Agent automatically synthesizes and validates executable evaluation pipelines. Extensive experiments show that A2Eval mitigates benchmark coverage imbalance, evaluation ranking distortion, and computational cost.





\section*{Impact Statement}
This paper presents work whose goal is to advance the field of machine learning, specifically in embodied vision-language model evaluation. Our proposed framework, A2Eval, enables more efficient, high-fidelity, and fair evaluations, reducing computational costs and mitigating systematic ranking biases. 
While we do not anticipate direct ethical risks or negative societal consequences, the reduction in computational resources may have a positive environmental impact by lowering energy consumption. Overall, our work is intended to support the research community by improving evaluation quality and efficiency, and does not introduce any known harmful effects.

\bibliography{paper}
\bibliographystyle{icml2025}
\newpage
\appendix

\begin{table*}[t]
\centering
\small
\setlength{\tabcolsep}{1.0pt}
\renewcommand{\arraystretch}{1}
\begin{tabular}{lccccccccc c}
\toprule
\textbf{Model} & PercepObj & SceneAct & SpatGeo & QuantNum & AffdFunc & PhysCaus & DecPlan & DynScene & \textbf{Avg} \\
\midrule
Qwen3-VL-235B-A22B-Thinking    & 62.23 & 60.03 & 71.23 & 65.22 & 68.58 & 65.68 & 68.26 & 66.56 & 65.97 \\
Internvl-3.5-241B-A28B         & 61.45 & 62.61 & 70.99 & 60.23 & 62.30 & 67.56 & 71.58 & 68.68 & 65.68 \\
GPT-5-20250807-Mini            & 64.80 & 65.23 & 72.17 & 61.90 & 73.22 & 60.58 & 63.20 & 63.07 & 65.52 \\
Qwen3-VL-30B-A3B-Thinking      & 61.04 & 57.62 & 68.87 & 59.77 & 65.03 & 61.40 & 64.44 & 62.48 & 62.58 \\
Internvl-3.5-38B               & 55.81 & 58.68 & 67.96 & 58.91 & 64.55 & 64.86 & 67.62 & 60.69 & 62.37 \\
Qwen3-VL-235B-A22B-Instruct    & 64.44 & 62.69 & 54.95 & 57.90 & 66.39 & 62.72 & 65.08 & 60.90 & 61.88 \\
Internvl-3.5-30B-A3B           & 54.08 & 58.67 & 69.81 & 57.84 & 54.10 & 66.36 & 68.10 & 65.37 & 61.79 \\
Qwen3-VL-30B-A3B-Instruct      & 55.03 & 53.23 & 67.45 & 58.26 & 60.11 & 67.10 & 66.30 & 64.30 & 61.47 \\
Internvl-3.5-8B                & 51.09 & 50.62 & 61.79 & 53.36 & 55.74 & 58.22 & 59.24 & 57.80 & 55.98 \\
Qwen2.5-VL-72B-Instruct        & 57.48 & 56.03 & 49.06 & 51.93 & 67.21 & 45.60 & 55.16 & 55.61 & 54.76 \\
Qwen2.5-VL-32B-Instruct        & 53.64 & 52.80 & 45.52 & 46.14 & 56.83 & 39.72 & 57.46 & 54.08 & 50.77 \\
Qwen2.5-VL-7B-Instruct         & 49.05 & 51.02 & 43.16 & 47.42 & 52.46 & 40.52 & 50.84 & 51.30 & 48.22 \\
Qwen2.5-VL-3B-Instruct         & 42.84 & 48.44 & 36.56 & 33.69 & 48.09 & 22.60 & 41.06 & 41.87 & 39.39 \\
\bottomrule
\end{tabular}
\caption{Performance of evaluated models across all capability dimensions. All scores are percentages. ``Avg'' is the mean across dimensions.}
\label{tab:model_scores}
\end{table*}

\section{Benchmark Description and Redundancy Analysis}
\label{app:benchmark}

This section provides supplementary information about the source benchmarks and illustrates the redundancy patterns in the evaluation landscape.

\textbf{Benchmark overview.} Table~\ref{tab:benchmark_info} summarizes all benchmarks used in our study, including their descriptions and references.

\begin{figure}[t!]
  \centering
  \includegraphics[width=0.9\columnwidth]{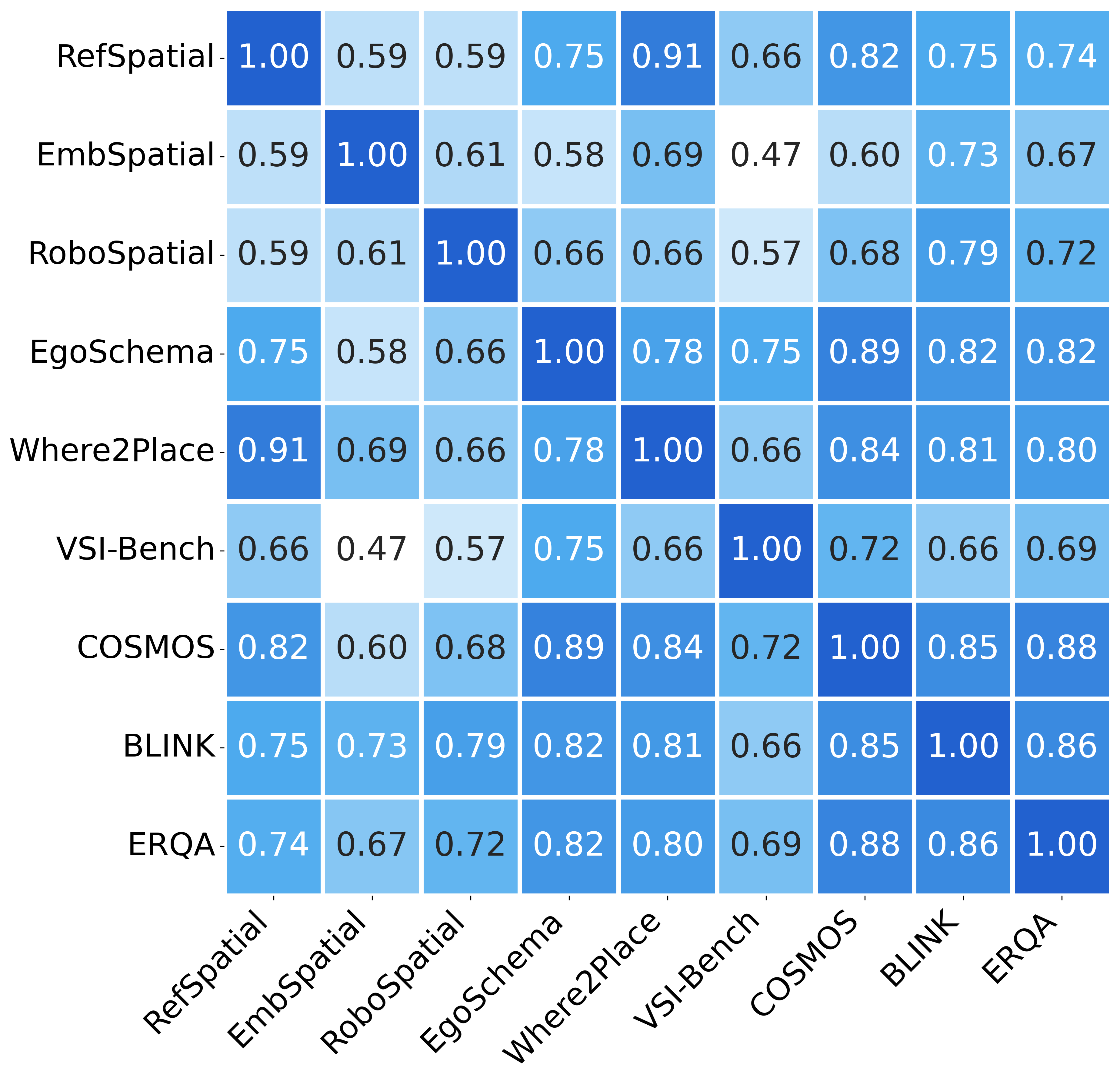}
  \caption{Heatmap of inter-benchmark similarity. The results reveal severe redundancy in the current evaluation ecosystem, with benchmarks exhibiting up to \textit{91\%} overlap (e.g., RefSpatial vs. Where2Place).}
  \label{fig:bmk_heatmap}
\end{figure}

\textbf{Inter-benchmark redundancy.} Figure~\ref{fig:bmk_heatmap} shows pairwise semantic similarity across benchmarks, revealing substantial overlap in the current evaluation ecosystem. Table~\ref{tab:redundancy_examples} provides concrete examples of redundant samples across different benchmarks.

\begin{figure}[ht!]
  \centering
  \includegraphics[width=\columnwidth]{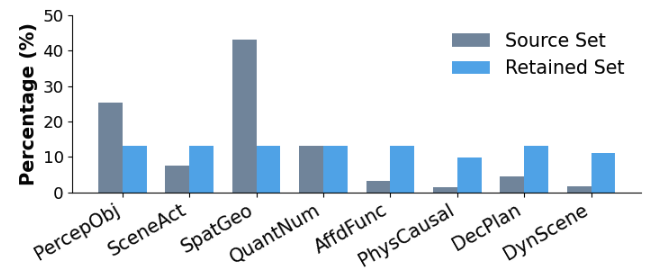}
  \caption{Capability distribution across dimensions before and after diversity-aware sampling. The retained set exhibits a substantially more balanced coverage compared to the highly skewed source distribution.}
  \label{fig:Cap Dist}
\end{figure}

\begin{figure*}[!htbp]
\centering

\begin{minipage}{0.24\textwidth}
\centering
\includegraphics[width=\linewidth]{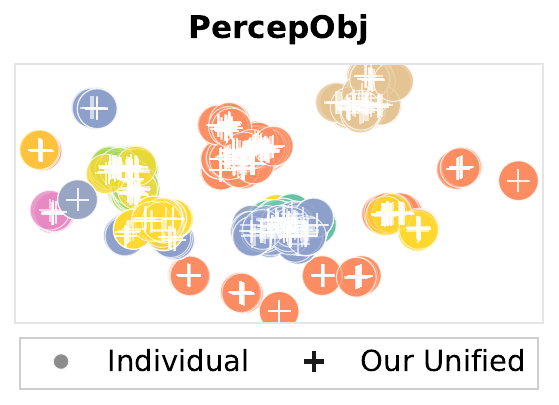}
\label{fig:percepobj}
\end{minipage}\hfill
\begin{minipage}{0.24\textwidth}
\centering
\includegraphics[width=\linewidth]{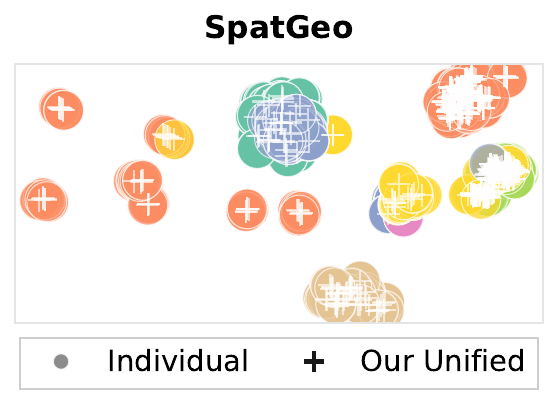}
\label{fig:spatgeo}
\end{minipage}\hfill
\begin{minipage}{0.24\textwidth}
\centering
\includegraphics[width=\linewidth]{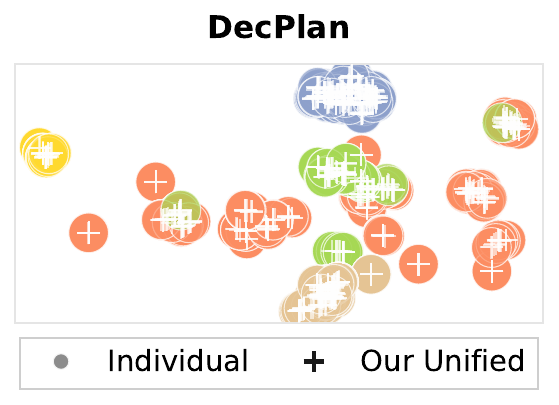}
\label{fig:decplan}
\end{minipage}\hfill
\begin{minipage}{0.24\textwidth}
\centering
\includegraphics[width=\linewidth]{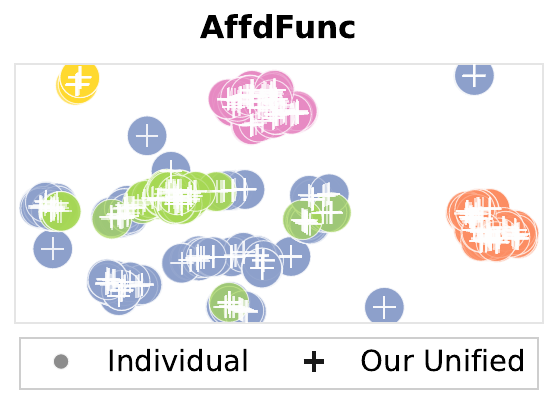}
\label{fig:affdfunc}
\end{minipage}

\vspace{0.8em}

\begin{minipage}{0.24\textwidth}
\centering
\includegraphics[width=\linewidth]{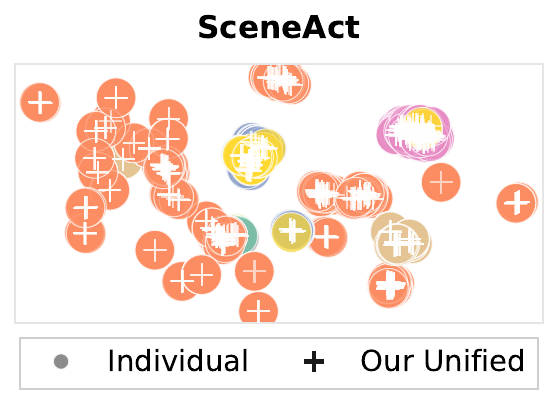}
\label{fig:sceneact}
\end{minipage}\hfill
\begin{minipage}{0.24\textwidth}
\centering
\includegraphics[width=\linewidth]{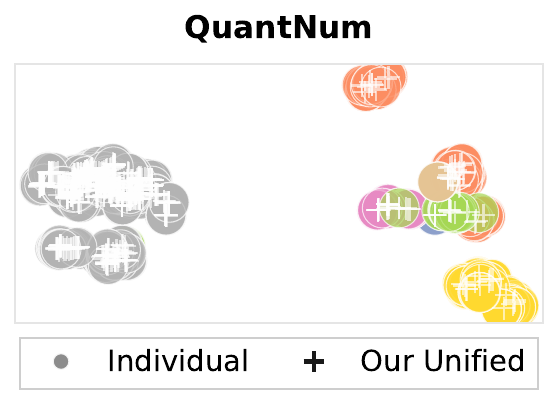}
\label{fig:quantnum}
\end{minipage}\hfill
\begin{minipage}{0.24\textwidth}
\centering
\includegraphics[width=\linewidth]{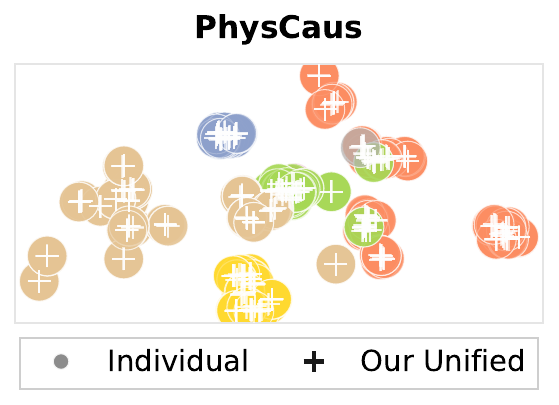}
\label{fig:physcaus}
\end{minipage}\hfill
\begin{minipage}{0.24\textwidth}
\centering
\includegraphics[width=\linewidth]{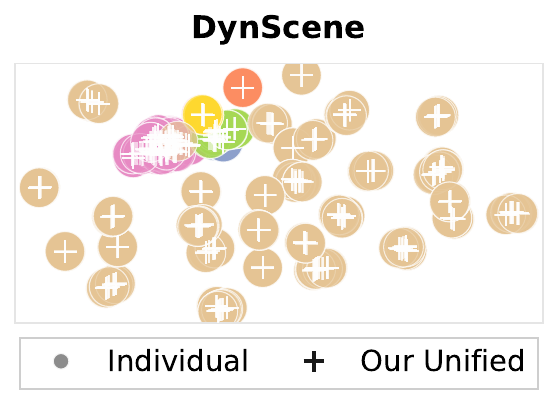}
\label{fig:dynscene}
\end{minipage}

\caption{UMAP visualization of sample embeddings across all dimensions.}
\label{fig:reasoning_modules}
\end{figure*}

\textbf{Dimension distribution and coverage.} Figure~\ref{fig:Cap Dist} shows the capability distribution before and after our sampling approach. Figure~\ref{fig:reasoning_modules} provides UMAP visualizations of sample embeddings for each dimension, with our selected samples marked by "+".

\begin{table}[h]
\centering
\footnotesize
\setlength{\tabcolsep}{4pt}
\renewcommand{\arraystretch}{1.12}
\begin{tabularx}{\columnwidth}{p{1.6cm}X}
\toprule
\textbf{Benchmark} & \textbf{Example} \\
\midrule
COSMOS & The robot is instructed to 'close the drawer.' [...] what is the most plausible next action? (A) move down [...] \\
ERQA & What action should the robot take to close the drawer? A. Move forward [...] \\
\midrule
RefSpatial-Bench & Point out the free spot behind the pink cup with equal distance to the red bowl. Your answer should be formatted [...] \\
Where2Place & Find spots between the blue mug and orange bowl. Your answer should be formatted [...] \\
\midrule
VSIBench & How many towel(s) are in this room? Please answer the question [...] \\
ERQA & How many unique towels and/or mats are in the room? A. two [...] \\
\midrule
VSIBench & Which object (chair, backpack, window, clock) is closest to the door? A. chair [...] \\
EmbSpatial-Bench & Which object is at the shortest distance? A. chair [...] \\
\bottomrule
\end{tabularx}
\caption{Four pairs of redundant samples across benchmarks.}
\label{tab:redundancy_examples}
\end{table}

\section{Detailed Evaluation Setting}
\label{app:eval_setting}

We provide the detailed evaluation settings for different model families below. All models are evaluated on $A800$

\paragraph{Qwen3-VL Models.}
Following the official GitHub recommendations, for instruct models, we configure Qwen3-VL Instruct models with $\text{top-p}=0.8$, $\text{top-k}=20$, $\text{temperature}=0.7$, $\text{repetition\_penalty}=1.0$. For thinking models, we use: $\text{top-p}=0.95$, $\text{top-k}=20$, $\text{temperature}=0.6$, $\text{repetition\_penalty}=1.0$.  $\text{min\_pixels}=2{,}097{,}152$, $\text{max\_pixels}=6{,}291{,}456$, and $\text{total\_pixels}=128000 * 32 * 32 * 0.9$ are used for all models

\paragraph{Qwen2.5-VL and InternVL-3.5 Models.}
For Qwen2.5-VL and InternVL-3.5 series, we use greedy decoding with temperature set to 0.01.
Note that InternVL-3.5-241B-A28B is deployed with the PyTorch backend.

\section{Implementation Details of the Evaluation Agents}
\label{app:pipeline}

\subsection{Overview of the Evaluation Pipeline}

This appendix provides implementation details of the two cooperative
agents used in our framework: (i) the \textit{Data Agent}, which
constructs capability dimensions and balanced benchmark with an Assigner equipped with 5 voter agents, and
(ii) the \textit{Eval Agent}, which synthesizes executable inference and
scoring pipelines for model evaluation.
We present full algorithmic specifications and the prompts used by
different agent roles.

\subsection{Data Agent: Dimension Construction and Benchmark Construction}
\label{app:data_agent}
\subsubsection{Algorithmic Overview}
The Data Agent constructs capability dimensions and builds a structured
benchmark through iterative proposal–review–assignment and diversity-aware
sampling. The full procedure is given in
Algorithm~\ref{alg:data_agent}.
\subsubsection{Roles of the Data Agent}
\paragraph{Proposer $\mathcal{A}_p$}
Generates candidate capability dimensions $D$ based on benchmark metadata
$\mathcal{B}$, current critique $C$ from the Reviewer, and historical memory
$M$ that records previous proposals and feedback. By leveraging $M$, the 
Proposer avoids repeating previously rejected dimension configurations 
(e.g., redundant merges or incomplete decompositions) and refines dimensions 
according to concrete suggestions in $C$.

\paragraph{Reviewer $\mathcal{A}_r$}
Evaluates candidate dimensions $D$ for conceptual redundancy, coverage
completeness, and data balance using balance statistics $S$ from the Assigner. 
Provides structured critique $C$ to guide dimension refinement, such as 
suggesting merges for redundant categories or introducing missing aspects. 
When $S$ reveals severe imbalances (e.g., $|E_{d_k}^{\text{s}}|$ too small for some dimension 
$d_k$), the Reviewer incorporates this information into its critique to trigger 
dimension adjustments in the next iteration.

\paragraph{Assigner $\mathcal{A}_a$ (with Voters $\{\mathcal{A}_a^i\}_{i=1}^{N_v}$)}
Performs two key operations for each candidate dimension set $D$: 
\textbf{(i) Dimension assignment} through an ensemble of $N_v$ voters that 
independently label each example $e_j \in \mathcal{E}$ with dimensions from $D$, 
with the final assignment determined by majority vote; 
\textbf{(ii) Diversity-aware sampling} that clusters examples within each 
dimension $d_k$ using embedding-based clustering and selects representatives 
from each cluster to maximize semantic diversity while removing intra-dimension 
redundancy. 
Computes balance statistics $S = \{|E_{d_k}^{\text{s}}|\}_{k=1}^{|D|}$ to identify 
under-represented dimensions and reports them to the Reviewer for potential 
merging or refinement.

\subsubsection{Prompt Design for the Data Agent}
\label{app:data_prompts}
\begin{promptbox}{The Prompt of Proposer}
\begin{lstlisting}
You are given descriptions of several benchmarks that evaluate various capabilities of vision-language models (VLMs).
Your task is to synthesize these sources and propose a new, unified set of evaluation dimensions specifically tailored for VLM evaluation for embody agent domain.

## Information Provided

[Start of Benchmark Description]
${description}
[End of Benchmark Description]

[Start of Historical Memory]
{memory}
[End of Historical Memory]

[Start of Current Critique]
{critique}
[End of Current Critique]

## Requirements

1. Dimensions must be non-overlapping, mutually independent, rationality.
2. Dimension name must be concise, capturing only the core capability of the dimension, with all non-essential qualifiers removed.
3. Dimension descriptions must be concise, precise, and represent a generalizable capability.
4. The historical memory section(optional) records previous dimension proposals and their critiques; use this information to avoid repeating rejected configurations or previously identified issues.
5. The current critique section is optional; if present, you must incorporate the critique feedback by adjusting, merging, refining, or splitting dimensions, ensuring the final set resolves any highlighted weaknesses or overlaps.
6. Output must be in clean JSON format with no additional commentary.

{{
    "name": "DimensionName",
    "description": "A one- or two-sentence explanation of the dimension."
}}

\end{lstlisting}
\end{promptbox}

\begin{promptbox}{The Prompt of Reviewer}
\begin{lstlisting}

You are a focused critical evaluator tasked with assessing the provided dimension framework for evaluating Vision-Language Models (VLMs) in a specific domain.

## Information Provided

[Start of Domain Description]
{domain}
[End of Domain Description]

[Start of Current Framework] 
{dimension_desc}
[End of Current Framework]

[Start of Balance Statistics]
{balance_stats}
[End of Balance Statistics]

## Requirements

1. Provide a textual evaluation of the current framework based on the following criteria, incorporating feedback from the balance statistics if provided.
   a. Specificity: Are the dimensions domain-specific rather than general model properties (e.g., do NOT include robustness, efficiency, handling unseen objects or scenes, end-to-end capability, or generalization)?
   b. Minimality: Does the framework avoid unnecessary fragmentation? Identify dimensions that should be merged due to strong dependency, lack of standalone evaluability, or conceptual overlap.
   c. Balance: If balance statistics are provided (e.g., dimension frequency counts), assess whether certain dimensions have too few examples and should be merged with related dimensions to ensure adequate coverage.
2. Return only a JSON object in the following format, with textual feedback for each criterion:

{{
    "specificity_feedback": "",
    "minimality_feedback": "",
    "balance_feedback": ""
}}

\end{lstlisting}
\end{promptbox}

\begin{promptbox}{The Prompt of Voters}
\begin{lstlisting}
You are a domain expert in a specific field. Your task is to assign a new sample to the single most relevant predefined dimension.

## Information Provided

You will receive the target domain description, the current set of predefined evaluation dimensions and the sample itself.

[Start of Domain Description]
{domain}
[End of Domain Description]

[Start of Predefined Dimensions]
{dimensions}
[End of Predefined Dimensions]

[Start of New Example]
{example}
[End of New Example]

## Requirements

1. Based on the dimension names and descriptions, determine the single most relevant dimension that best captures the primary capability being evaluated in the sample.
2. If the sample does not match any dimension well, assign it to "other".
3. Return a JSON object in the following exact format:

{{
    "name": "dimension_name_or_other",
    "reason": "explain why the sample fits this dimension"
}}

\end{lstlisting}
\end{promptbox}


\subsection{Eval Agent: Model Inference and Scoring Logic Synthesis}
\label{app:eval_agent}
\subsubsection{Algorithmic Overview}
The Eval Agent synthesizes executable inference logic and scoring logic
through feedback-guided refinement with a sandbox executor. The complete
procedure is shown in Algorithm~\ref{alg:eval_agent}.

\subsubsection{Roles of the Eval Agent}
\paragraph{Evaluator $\mathcal{A}_e$}
Synthesizes inference logic $F^e$ that loads test inputs from $\mathcal{T}$,
executes the target model $M$ under the required evaluation protocol, and
produces predictions $\mathcal{P}$. Iteratively refines $F^e$ based on
diagnostic feedback from the sandbox executor $\mathcal{X}$ until execution
is reliable.

\paragraph{Scorer $\mathcal{A}_s$}
Synthesizes scoring logic $F^s$ that computes evaluation metrics based on
test cases $\mathcal{T}$, model predictions $\mathcal{P}$, and benchmark
specifications. Iteratively refines $F^s$ based on feedback from $\mathcal{X}$
to ensure metric consistency and correctness.

\paragraph{Sandbox Executor $\mathcal{X}$}
Executes generated code ($F^e$ and $F^s$) in an isolated environment and
returns diagnostic feedback (e.g., runtime errors, API mismatches) when
execution fails, enabling iterative refinement by the Evaluator and Scorer.

\subsubsection{Prompt Design for the Eval Agent}
\label{app:eval_prompts}

\begin{promptbox}{The Prompt of Evaluator}
\begin{lstlisting}

You will write a VLM inference function based on the provided VLM model type, benchmark example, and optional additional information.
Your output must be a complete, self-contained Python function that performs VLM inference using the given model path, benchmark inputs, model generation parameters, and visual parameters.

## The Information Provided to You
[Start of Model Type]
{vlm_type}
[End of Model Type]
[Start of Benchmark Example]
{benchmark_example}
[End of Benchmark Example]

## Optional Additional Information
This section may or may not be present. If present, use it to improve or fix your generated code.
It contains:
1. The previously generated code result.
2. The execution error message produced when running that code on the benchmark examples.
[Start of Previous Generated Code]
{previous_code}
[End of Previous Generated Code]
[Start of Execution Error]
{execution_error}
[End of Execution Error]

## Task Overview
Your goal is to produce a reliable and fully functional VLM inference function that:
1. Loads the specified VLM model only once from the given model path before processing any samples.
2. Accepts benchmark inputs as a list of dicts, each containing a question and visual inputs (image or video), and supports multi-sample inference.
3. Handles model generation parameters provided through gen_kwargs (default: max_new_tokens 1024, temperature 0.01, seed 42).
4. Applies visual parameters from visual_kwargs (default: max_frames 32). Where max_frames defines the maximum number of frames to sample from a video. If the video contains fewer than max_frames frames, all frames should be used; if it contains more, it should be uniformly sampled down to max_frames frames.
5. Performs inference for each sample and returns only the model's predicted outputs as a list of strings.
6. Uses the following function signature:
   def run_vlm_inference(model_path: str, benchmark_input: list[dict], gen_kwargs: dict = {{ "max_new_tokens": 1024, "temperature": 0.01, "seed": 42 }}, visual_kwargs: dict = {{ "max_frames": 32 }}) -> list[str]
7. If optional additional information is provided, use it to refine, debug, or correct your code generation.
8. Your code must not use any try/except blocks or other exception-handling constructs.
9. Must include a visible multi-sample progress bar (e.g., tqdm) during inference.
10. Enforces the following output format: produce a single Python code block with the implementation, with no explanations, comments, or text outside the code block.

\end{lstlisting}
\end{promptbox}

\begin{promptbox}{The Prompt of Scorer}
\begin{lstlisting}

You will write a benchmark scoring function based on the provided benchmark information, benchmark example, and optional additional information.
Your output must be a complete, self-contained Python function that computes a score using the benchmark's scoring methodology.

## The Information Provided to You
[Start of Benchmark Information]
{benchmark_infor}
[End of Benchmark Information]
[Start of Benchmark Example]
{benchmark_example}
[End of Benchmark Example]

## Optional Additional Information
This section may or may not be present. If present, use it to improve or fix your generated scoring logic.
It contains:
1. The previously generated scoring function.
2. The execution error message produced when running that function on the benchmark examples.
[Start of Previous Generated Code]
{previous_code}
[End of Previous Generated Code]
[Start of Execution Error]
{execution_error}
[End of Execution Error]

## Task Overview
Your goal is to produce a reliable and fully functional scoring function that:
1. Infers the correct scoring rules of the benchmark based on the benchmark name and example.
2. Correctly handles different evaluation types (e.g., exact match questions, numerical questions, multiple-choice, or any benchmark-specific formats).
3. Computes and returns a float score for a single prediction answer pair.
4. Uses the following function signature exactly:
   def eval_score(pred, answer) -> int
5. Ensures that the function is self-contained, does not rely on external files, and does not use any try/except blocks or other exception-handling constructs.
6. If optional additional information is provided, use it to refine, debug, or correct your generated scoring function.
7. Enforces the following output format: produce a single Python code block with the implementation, with no explanations, comments, or text outside the code block.

\end{lstlisting}
\end{promptbox}

\section{Human--Agent Agreement Protocol}
\label{app:human_agent_agreement}

This appendix details the human-in-the-loop verification protocol used to validate that the Assigner's automatic dimension assignments align with human expert understanding.

\subsection{Annotation Design}
Rather than asking annotators to assign dimensions from scratch---which would require full familiarity with the entire taxonomy and impose substantial cognitive burden---we adopt a two-stage annotation process that combines strong model pre-annotation with expert human verification.

\textbf{Stage 1: Model Pre-Annotation.} A strong reference model (GPT-4o) performs preliminary dimension annotation for each sample by proposing the most relevant capability dimensions based on the full taxonomy definition.

\textbf{Stage 2: Human Verification.} Human annotators review these model-proposed labels, correct errors when necessary, and produce the final validated dimension assignments. This design significantly reduces annotator cognitive load while preserving strict human oversight, allowing annotators to focus on judgment refinement and boundary cases rather than taxonomy recall.

\subsection{Annotator Recruitment}
Human annotation is performed by 5 domain experts, including PhD researchers and engineers with relevant backgrounds in embodied intelligence. All annotators undergo a unified calibration process with shared dimension definitions and representative examples prior to annotation, ensuring consistent interpretation of capability boundaries. During annotation, model predictions are used only as references; annotators make independent decisions and revise model outputs whenever ambiguity or clear misclassification is observed.

\subsection{Annotation Procedure}
For each dimension, we construct a balanced evaluation set consisting of candidate positive and candidate negative samples proposed by the Data Agent. All samples are fully randomized and presented in a blind manner: annotators are not informed of the Data Agent's predictions, benchmark sources, or class balance. Each sample is independently reviewed by multiple annotators, and final dimension labels are determined via majority voting. For each dimension, we sample 100 examples for evaluation.

\subsection{Agreement Metrics}
We measure three types of agreement:
\begin{itemize}[leftmargin=*,itemsep=2pt]
\item \textbf{Cohen's $\kappa$}: Agreement beyond chance between automatic assignments and human-validated labels, accounting for random agreement.
\item \textbf{Inter-Annotator Agreement (IAA)}: Fleiss' $\kappa$ among the five expert annotators, measuring consistency of human judgment.
\end{itemize}

\subsection{Results and Analysis}

\begin{table}[t]
\centering
\footnotesize
\setlength{\tabcolsep}{6pt}
\renewcommand{\arraystretch}{1.15}

\begin{tabular}{lcc}
\toprule
\textbf{Dimension} &
\textbf{Cohen's $\kappa$} &
\textbf{IAA} \\
\midrule
PercepObj              & 0.81          & 0.83          \\
SceneAct               & 0.82          & 0.84          \\
SpatGeo                & 0.79          & 0.80          \\
QuantNum               & 0.77          & 0.78          \\
DynScene               & 0.75          & 0.76          \\
AffdFunc               & 0.80          & 0.82          \\
PhysCausal             & 0.76          & 0.77          \\
DecPlan                & 0.78          & 0.79          \\
\midrule
\textbf{Average}       & \textbf{0.78} & \textbf{0.80} \\
\bottomrule
\end{tabular}

\caption{Human--agent agreement on capability dimension assignment using a
human-in-the-loop verification protocol.
Cohen's $\kappa$ measures agreement between automatic assignments and
human-validated labels beyond chance, while IAA reports inter-annotator
agreement among the five expert annotators.}
\label{tab:human_agent_agreement}
\end{table}

Table~\ref{tab:human_agent_agreement} summarizes the human--agent agreement results. Across all dimensions, we observe strong agreement between automatic dimension assignments and human-validated labels, with an average Cohen's $\kappa$ of 0.78 and an average inter-annotator agreement of 0.80, indicating substantial to near-perfect consistency. These results demonstrate that the automatically induced capability dimensions are well aligned with expert human judgment.

\section{Human-Centric Evaluation Protocol}
\label{app:human_eval_protocol}

This appendix details the human-centric evaluation protocol.
The goal is to construct a human-perceived ranking of embodied planners and
compare it with the ranking induced by our new benchmark, testing whether the
benchmark signal aligns with human intuition about model capability in realistic
embodied scenarios.

\subsection{Objective}

We build a human-perceived evaluation signal as a complementary axis to automated
benchmarking: annotators watch real-world videos and judge the quality of models'
embodied action plans. We then compare the resulting human-centric model ranking
against the new-benchmark ranking to verify (i) whether the new benchmark reflects
human-perceived capability differences and (ii) whether the induced capability
dimensions better disentangle model strengths with reduced redundancy.

\subsection{Scenarios}

We use eight real-world video scenarios: \textbf{Household}, \textbf{Grocery Store}, \textbf{School}, \textbf{Office}, \textbf{Warehouse}, \textbf{Hospital}, \textbf{Workshop}, and \textbf{Laboratory}. Each video lasts approximately 15--30 seconds, and each
scenario contains 4--6 sub-tasks described in a natural instruction.

\subsection{A Concrete Example: Household Scenario}

We include one concrete household example to illustrate the evaluation setup.
The household instruction is:

\begin{quote}
``I'm going to work. Please put the shoes onto the shoe rack, throw the trash on
the table into the trash bin, and put the dirty clothes on the sofa into the washing
machine.''
\end{quote}

This instruction is decomposed into three sub-tasks:
(1) \textbf{Shoe organization}: identify shoes and the shoe rack, navigate to each shoe,
grasp and place shoes onto appropriate rack slots while keeping pairs and orientations
reasonable;
(2) \textbf{Table trash disposal}: identify the table and distinguish trash from non-trash
items, navigate to the table, grasp trash items one-by-one, navigate to the trash bin,
and deposit them until no visible trash remains;
(3) \textbf{Sofa laundry loading}: identify the sofa region and dirty clothes, navigate and
grasp clothes, locate the washing machine and its opening direction, navigate to the
machine, open the door, place clothes into the drum, and close the door, repeating
until the sofa is cleared.

This example is shown to models as a short video clip (15--30s) together with the
instruction, and models are asked to produce an explicit step-by-step plan that is
physically feasible and consistent with what is visible.

\subsection{Model Input and Required Output Format}

All models receive the same input consisting of:
(i) the scenario video, (ii) a global system prompt, and (iii) a user prompt containing
the task instruction.
We enforce a unified output structure for all models to ensure comparability:

\begin{promptbox}{System Prompt (global).}
You are an embodied AI planner.
You will receive a short video that shows the current environment.
Your task is to understand the scene, identify relevant objects,
and generate a complete and physically feasible action plan to achieve the goal.
You must rely ONLY on what is visible or inferable from the video.
Do NOT assume the presence of objects or locations that are not shown in the video.
Do NOT skip necessary physical steps (e.g., opening containers, moving closer, reaching).
Follow the required output structure strictly.
\end{promptbox}

\begin{promptbox}{User Prompt}
[Environment Video Provided]\\
Please generate an explicit, step-by-step embodied action plan based solely on
what is visible in the video.

Your response must make your reasoning explicit with respect to:
(1) scene elements and object locations,
(2) task goals and sub-goals,
(3) action order and temporal dependencies,
(4) spatial relations and navigation trajectories,
(5) object states and state changes over time,
(6) quantities and counts of relevant objects whenever applicable.

Please structure your output as follows:\\
1. Scene Understanding\\
2. Goal Interpretation\\
3. Action Plan (Step-by-step)\\
4. Final Plan (condensed format)
\end{promptbox}

\subsection{Human Rating Dimensions and Weights}

Annotators score each model output based on the video and the model's response.
We use \textbf{eight} criteria, each rated on a 1--5 Likert scale, and aggregated
with \textbf{equal weights (12.5\% each)}:

\begin{itemize}
    \item \textbf{Planning}: whether the plan covers key steps and completes the instruction;
    \item \textbf{Scene Understanding}: whether the model correctly perceives and describes the scene structure;
    \item \textbf{Dynamic Consistency}: whether the plan remains consistent with object states/locations over time;
    \item \textbf{Object Grounding}: whether referenced target objects match what is in the video (e.g., the correct item);
    \item \textbf{Physical Reasoning}: whether actions are physically feasible and respect object states;
    \item \textbf{Quantitative Reasoning}: whether counting/amount-related decisions are correct (e.g., number of items);
    \item \textbf{Spatial Reasoning}: whether spatial relations and navigation/trajectory are used correctly;
    \item \textbf{Temporal Reasoning}: whether the action order is sensible and respects prerequisite constraints.
\end{itemize}

\subsection{Human Evaluation Procedure}

For each scenario, the annotators (1) watch the video, (2) read the instruction, (3) read the anonymized model output, and (4) score each output along the eight dimensions. Each model-scenario instance is rated independently by at least 3--5 annotators.
Both scenario and model order are randomized to mitigate ordering bias.
Table~\ref{tab:household_dimension_example} provides a concrete, dimension-wise example of the human evaluation rubric and corresponding model outputs for a representative household scenario.

\subsection{Score Aggregation and Ranking Alignment}

For each model output, we compute a weighted overall score by averaging the
eight dimension scores (each normalized to the same scale and weighted equally).
We then average scores across annotators and across all scenarios to obtain a
final human-centric score per model, which induces a human-perceived ranking.

In parallel, we compute the benchmark score and ranking for the same model set
under our new benchmark. We quantify ranking alignment between benchmark-induced
and human-centric rankings using Spearman's $\rho$ and Kendall's $\tau$.
High correlation indicates that the new benchmark captures evaluation signals
consistent with human judgments, and supports the claim that the induced capability
dimensions reflect human-perceived embodied competence.

\section{Detailed Fidelity Analysis}
\label{appendix:fidelity}

This section provides detailed breakdowns of the Eval Agent's fidelity evaluation presented in Section~\ref{subsec:eval_agent_fidelity}. We separately analyze inference fidelity and scoring fidelity.

\begin{table}[!htbp]
\centering
\footnotesize
\setlength{\tabcolsep}{3pt}
\renewcommand{\arraystretch}{1.12}
\begin{tabular}{lcccc}
\toprule
\textbf{Dimension} &
\textbf{Q2.5-7B} &
\textbf{Q3-8B} &
\textbf{IVL3.5-8B} &
\textbf{Avg} \\
\midrule
PercepObj  & 92.5 & 92.4 & 96.5 & 93.8 \\
SceneAct   & 92.0 & 91.3 & 92.5 & 91.9 \\
SpatGeo    & 94.5 & 93.4 & 94.7 & 94.2 \\
QuantNum   & 94.9 & 93.8 & 94.4 & 94.4 \\
AffdFunc   & 91.8 & 91.1 & 92.3 & 92.7 \\
PhysCaus   & 91.6 & 91.9 & 92.4 & 92.0 \\
DecPlan    & 96.6 & 95.6 & 96.9 & 96.4 \\
DynScene   & 93.9 & 93.4 & 93.7 & 93.6 \\
\midrule
\textbf{Average} & 93.5 & 92.9 & 94.2 & \textbf{93.6} \\
\bottomrule
\end{tabular}
\caption{Inference fidelity between agent-generated inference outputs and reference model results.}
\label{tab:inference_fidelity}
\end{table}

\subsection{Inference Fidelity}
Table~\ref{tab:inference_fidelity} reports the fidelity of inference logic $F^f$ generated by the Eval Agent. We measure fidelity using normalized edit distance between agent-generated outputs and reference model results: $\text{Fidelity}_{\text{inf}} = 1 - \text{EditDist}_{\text{norm}}$, where the edit distance is normalized to $[0,1]$ by dividing by the maximum possible distance (the length of the longer sequence). The agent achieves 93.6\% average inference fidelity across all capability dimensions and model configurations.

\subsection{Scoring Fidelity}
Table~\ref{tab:scoring_fidelity} presents the fidelity of scoring logic $F^s$ generated by the Eval Agent. We measure fidelity using relative score difference: $\text{Fidelity}_{\text{score}} = 1 - \frac{|\text{Score}_{\text{pred}} - \text{Score}_{\text{ref}}|}{|\text{Score}_{\text{ref}}|}$, where $\text{Score}_{\text{pred}}$ and $\text{Score}_{\text{ref}}$ denote the scores computed by agent-generated and reference scoring functions, respectively.

The scoring fidelity exceeds 97\% on average across all capability dimensions and model configurations. Rare discrepancies stem from borderline cases near numerical thresholds rather than fundamental errors in the generated scoring logic.

\begin{table}[!htbp]
\centering
\footnotesize
\setlength{\tabcolsep}{3pt}
\renewcommand{\arraystretch}{1.12}
\begin{tabular}{lcccc}
\toprule
\textbf{Dimension} &
\textbf{Q2.5-7B} &
\textbf{Q3-8B} &
\textbf{IVL3.5-8B} &
\textbf{Avg} \\
\midrule
PercepObj  & 94.9  & 94.5 & 93.5  & 94.3 \\
SceneAct   & 98.4  & 98.5 & 99.0  & 98.6 \\
SpatGeo    & 95.8  & 97.9 & 96.5  & 96.7 \\
QuantNum   & 95.1  & 94.0 & 95.7  & 94.9 \\
AffdFunc   & 99.6  & 97.0 & 100.0 & 98.9 \\
PhysCaus   & 100.0 & 99.1 & 100.0 & 99.7 \\
DecPlan    & 100.0 & 98.3 & 100.0 & 99.4 \\
DynScene   & 100.0 & 96.4 & 100.0 & 98.8 \\
\midrule
\textbf{Average} & 97.9 & 97.0 & 97.9 & \textbf{97.9} \\
\bottomrule
\end{tabular}
\caption{Scoring fidelity between agent-generated and reference scoring functions.}
\label{tab:scoring_fidelity}
\end{table}

\begin{table*}[htbp]
\centering
\small
\setlength{\tabcolsep}{9pt}
\renewcommand{\arraystretch}{1.20}
\begin{tabular}{p{0.28\textwidth} p{0.58\textwidth}}
\toprule
\textbf{Dimension} & \textbf{Description} \\
\midrule

VSI-Bench \cite{yang2025thinking} &
VSI-Bench includes eight tasks of three types: configurational, measurement estimation, and spatiotemporal.
The configurational tasks (object count, relative distance, relative direction, route plan) test a model’s understanding of the configuration of a space and are more intuitive for humans. \\

EgoSchema \cite{mangalam2023egoschema} &
EgoSchema, a diagnostic benchmark for assessing very long-form video-language understanding capabilities of modern multimodal systems. 
Understanding long natural videos requires a host of interconnected abilities such as action and scene understanding, perceiving and tracking object states, long-term visual memory, abstract reasoning, hierarchical information aggregation, and more. \\

COSMOS \cite{azzolini2025cosmos} &
COSMOS focus on
(1) “task-completion verification”: the ability to determine whether a task or subtask has been successfully completed; 
(2) “action affordance”: the ability to assess whether executing a specific action or making progress toward a goal is possible;
(3) “next plausible action prediction”: the ability to identify the most plausible next action or subtask to advance toward a specified goal. \\

Where2Place \cite{yuan2025robopoint} &
Where2place evaluate free space reference using spatial relations. 
The images are collected from various cluttered environments. 
Each image is labeled with a sentence describing the desired some free space and a mask of the desired region. \\

ERQA \cite{team2025gemini} &
ERQA, a benchmark that focuses specifically on capabilities likely required by an embodied agent interacting with the physical world. 
ERQA consists of 400 multiple choice Visual Question Answering (VQA)-style questions across a wide variety of categories, 
including spatial reasoning, trajectory reasoning, action reasoning, state estimation, pointing, multi-view reasoning, and task reasoning. \\

EmbSpatialBench \cite{du2024embspatial} &
EmbSpatialBench, a benchmark for evaluating spatial understanding abilities of LVLMs in embodied environments.
Focus on six spatial relationships described from the egocentric perspective, including above, below, left,right, close and far, which completely covers three dimensions of the coordinates. \\

OmniSpatial \cite{jia2025omnispatial} &
OmniSpatial provides representative examples across its four main categories of spatial reasoning.
Perspective Taking: Tasks demonstrate the ability to understand spatial relationships from different viewpoints.
Dynamic Reasoning: This includes tasks that involve understanding object movement and changes.
Spatial Interaction: These tasks focus on engaging with spatial environments
Complex Logic: This category covers higher-level reasoning like pattern recognition (style, quantity, attributes, location) and geometric reasoning (polyhedron unfolding, sections and projections, mental rotation, assembly, analytical geometry). \\

RefSpatialBench \cite{zhou2025roborefer} &
RefSpatialBench comprising 200 real-world images with manually annotated tasks for object location and placement.
Fills the gap in evaluating spatial referring with multi-step reasoning. \\

RoboSpatial \cite{song2025robospatial} &
RoboSpatial: a new spatial reasoning benchmark designed to evaluate vision-language models (VLMs) in real-world indoor environments for robotics. 
It consists of 350 spatial reasoning questions paired with crowd-sourced RGBD images captured using a handheld iPhone camera equipped with a depth sensor. 
Each image is annotated with three types of spatial relationship questions—spatial configuration, spatial context, and spatial compatibility—providing a comprehensive evaluation of spatial understanding in robotic applications.
Spatial Configuration: Determines the relative positioning of objects (e.g., "Is the mug to the left of the laptop?").
Spatial Context: Identifies vacant areas in relation to a reference object (e.g., "Identify empty space to the left of the bowl.").
Spatial Compatibility: Assesses whether an object can fit within a specified area (e.g., "Can the chair be placed in front of the desk?"). \\

BLINK \cite{fu2024blink} &
BLINK evaluates a more comprehensive range of visual perception abilities, like multi-view reasoning, depth estimation, and reflectance estimation. \\
\bottomrule
\end{tabular}
\caption{Benchmarks with their description.}
\label{tab:benchmark_info}
\end{table*}

\begin{table*}[!htbp]
\centering
\small
\setlength{\tabcolsep}{9pt}
\renewcommand{\arraystretch}{1.25}
\begin{tabular}{p{3.2cm} p{5.8cm} p{6.0cm}}
\toprule
\textbf{Dimension} &
\textbf{Evaluation Rubric (Household Scenario)} &
\textbf{Example Model Output (qwen3-vl-235b)} \\
\midrule

Planning &
The plan fully completes all sub-tasks specified in the human instruction.
Specifically, does the plan include:
(1) placing \emph{all} shoes from the floor onto the shoe rack,
(2) disposing \emph{all} garbage items (e.g., paper cups and milk cartons) from the table into the trash bin,
and (3) transferring \emph{all} dirty clothes from the sofa into the washing machine,
followed by explicitly closing the washing machine lid? &
\emph{``Approach coffee table → Collect snack packages → Dispose in trash bin → Collect empty cups → Organize cups on table → Move to shoe area → Pick up floor shoes → Place on shoe rack → Verify tidy state... Approach the coffee table from the sofa area, moving approximately 1.5 meters forward to reach the table... Walk toward the shoe area near the white cabinet, moving approximately 3 meters across the room...''}
\\

Scene Understanding &
Does the plan correctly perceive and reference all key elements visible in the environment,
including the shoe rack on the left, the table in front of the sofa,
the trash bin, the sofa with dirty clothes, and the washing machine on the right,
without hallucinating unseen objects or assuming unsupported functionalities? &
\emph{``The video shows a modern living room with wooden flooring and several key elements:
- A white L-shaped sofa where a person in red clothing was seated
- A coffee table with multiple items: snack packages (chips), two white cups, a book, and a container
- A black wire trash bin positioned next to the coffee table
- A black metal shoe rack with 3-4 pairs of shoes (some on the rack, some on the floor)
- A washing machine near the gray curtains
...''}
\\

Dynamic Consistency &
Does the plan maintain consistent object states and locations throughout execution?
In particular, once shoes are placed onto the rack or clothes are moved into the washing machine,
are these objects no longer referred to as remaining in their original locations
in subsequent steps? &
\emph{``Pick up the empty snack packages from the coffee table surface (2-3 packages visible)... Move the snack packages to the trash bin located 0.5 meters to the left of the coffee table... Deposit the snack packages into the trash bin...''}
\\

Object Grounding &
Does the plan correctly ground actions to the intended target objects in the scene,
such as distinguishing trash items on the table from non-trash objects,
and identifying shoes on the floor versus clothes on the sofa,
without confusing them with nearby but irrelevant items? &
\emph{``Pick up the empty snack packages from the coffee table surface (2-3 packages visible)... Move the snack packages to the trash bin located 0.5 meters to the left of the coffee table...''}
\\

Physical Reasoning &
Does the plan respect physical constraints and affordances of objects,
such as opening the washing machine lid before loading clothes,
using appropriate body parts for grasping,
and avoiding physically implausible or unsafe actions? &
\emph{``Place the cups back on the coffee table in an organized manner (since no sink is visible in the scene)... Walk toward the shoe area near the white cabinet, moving approximately 3 meters across the room...''}
\\

Quantitative Reasoning &
Does the plan correctly reason about quantities in the scene,
such as how many shoes need to be picked up,
how many trash items (e.g., paper cups or milk cartons) are on the table,
and whether \emph{all} visible clothes on the sofa are transferred,
rather than treating multiple objects as a single aggregate action? &
\emph{``Pick up the two empty white cups from the coffee table... Pick up the shoes that are on the floor next to the shoe rack (3-4 pairs visible)...''}
\\

Spatial \& Trajectory &
Does the plan reason correctly about spatial relations and navigation trajectories,
including whether shoes are placed \emph{onto} the shoe rack (rather than merely near it),
and whether clothes are placed \emph{inside} the washing machine,
following plausible movement paths between locations? &
\emph{``Move the snack packages to the trash bin located 0.5 meters to the left of the coffee table...Approach the coffee table from the sofa area... Walk toward the shoe area near the white cabinet...''}
\\

Temporal Reasoning &
Does the plan respect temporal dependencies between actions,
such as navigating before grasping,
opening containers before use,
and completing one sub-task before transitioning to the next,
without interleaving unfinished steps? &
\emph{``A black metal shoe rack with 3-4 pairs of shoes (some on the rack, some on the floor)... Pick up the shoes that are on the floor next to the shoe rack... Place each pair of shoes on the appropriate shelf of the black metal shoe rack...''}
\\

\bottomrule
\end{tabular}

\caption{Dimension-wise human evaluation example for the household scenario.}
\label{tab:household_dimension_example}
\end{table*}

\begin{algorithm}[t]
\caption{Data Agent}
\label{alg:data_agent}
\begin{algorithmic}
\STATE \textbf{Input:} Benchmark information $\mathcal{B} = \{B_i\}_{i=1}^{N_B}$; 
Benchmark examples $\mathcal{E}=\bigcup_{i=1}^{N_B} E_i$
\STATE \textbf{Output:} Benchmark $\widetilde{\mathcal{B}}$ with dimensions $D$;
Dimension-assigned examples $\mathcal{E}^{\text{s}}$
\STATE \textbf{Roles:} Proposer $\mathcal{A}_p$; Reviewer $\mathcal{A}_r$; Assigner $\mathcal{A}_a$ (Voters $\{\mathcal{A}_a^1, \dots, \mathcal{A}_a^{N_v}\}$)
\STATE
\STATE \textbf{Initialize:} memory $M\gets[\;]$, critique $C_0\gets\texttt{""}$, balance stats $S^{(0)}\gets\emptyset$
\REPEAT
    \STATE \textbf{(a) Dimension Induction}
    \REPEAT
        \STATE $D_{i+1} \gets \mathcal{A}_p(\mathcal{B}, C_i, M)$
        \STATE $C_{i+1} \gets \mathcal{A}_r(D_{i+1}, S^{(r)})$
        \STATE $M \gets M \cup \{(D_{i+1}, C_{i+1})\}$
        \STATE $i \gets i + 1$
    \UNTIL{$D_{i+1} = D_{i}$}
    \STATE $D^{(r+1)} \gets D_i$
    \STATE
    \STATE \textbf{(b.1) Dimension Assignment}
    \FOR{each example $e_j \in \mathcal{E}$}
        \FOR{each voter $\mathcal{A}_a^k, k=1,\dots,N_v$}
            \STATE $D_{jk} \gets \mathcal{A}_a^k(e_j, D^{(r+1)})$
        \ENDFOR
        \STATE $D_j \gets \text{MajorityVote}(\{D_{jk}\}_{k=1}^{N_v})$
    \ENDFOR
    \FOR{each dimension $d_k \in D^{(r+1)}$}
        \STATE $E_{d_k} \gets \{e_j \mid d_k \in D_j \}$
    \ENDFOR
    \STATE
    \STATE \textbf{(b.2) Diversity-aware Sampling}
    \FOR{each dimension $d_k \in D^{(r+1)}$}
        \STATE $X_{d_k} \gets \{\text{Encoder}(e) : e \in E_{d_k}\}$
        \STATE Cluster $X_{d_k}$ into $K$ groups with centroids $\{\mu_j\}_{j=1}^K$
        \STATE $E_{d_k}^{\text{s}} \gets \big\{\arg\min_{x \in C_j} \|x-\mu_j\| : j=1,\dots,K \big\}$
    \ENDFOR
    \STATE $S^{(r+1)} \gets \{|E_{d_k}^{\text{s}}| : d_k \in D^{(r+1)}\}$
    \STATE $r \gets r + 1$
\UNTIL{$D^{(r+1)} = D^{(r)}$}
\STATE $D \gets D^{(r)}$
\STATE $\mathcal{E}^{\text{s}} \gets \bigcup_{k=1}^{|D|} E_{d_k}^{\text{s}}$
\end{algorithmic}
\end{algorithm}

\begin{algorithm}[t]
\caption{Eval Agent}
\label{alg:eval_agent}
\begin{algorithmic}
\STATE \textbf{Input:} Benchmark information $\mathcal{B}$; Test examples $\mathcal{T} \subseteq \mathcal{E}^{\text{s}}$;
Model $M$; Capability dimensions $D$; Benchmark examples $\mathcal{E}^{\text{s}} = \bigcup_{k=1}^{N_d} E_{d_k}^{\text{s}}$
\STATE \textbf{Output:} Inference logic $F^e$; Scoring logic $F^s$;
Model predictions $\{\mathcal{P}_{d_k}\}_{{d_k}\in D}$; Model scores $\{S_{d_k}\}_{{d_k}\in D}$
\STATE \textbf{Roles:} Evaluator $\mathcal{A}_e$; Scorer $\mathcal{A}_s$
\STATE \textbf{Sandbox:} Executor $\mathcal{X}$
\STATE
\STATE \textbf{Phase I: Model Inference Logic Construction}
\STATE $F^e_{0} \gets \mathcal{A}_e(M, \mathcal{T})$
\REPEAT
    \STATE $E_{i} \gets \mathcal{X}(F^e_i, M, \mathcal{T})$
    \STATE $F^e_{i+1} \gets \mathcal{A}_e(M, \mathcal{T}, F^e_{i}, E_i)$
    \STATE $i \gets i + 1$
\UNTIL{$E_i = \texttt{None}$ (no error feedback)}
\STATE $F^e \gets F^e_i$
\STATE
\STATE \textbf{Phase II: Model Scoring Logic Construction}
\STATE $\mathcal{P}_{\mathcal{T}} = F^e(M, \mathcal{T})$
\STATE $F^s_0 \gets \mathcal{A}_s(\mathcal{T}, \mathcal{P}_{\mathcal{T}}, \mathcal{B})$
\REPEAT
    \STATE $E_{j} \gets \mathcal{X}(F^s_j, \mathcal{T}, \mathcal{P}_{\mathcal{T}})$
    \STATE $F^s_{j+1} \gets \mathcal{A}_s(\mathcal{T}, \mathcal{P}_{\mathcal{T}}, \mathcal{B}, F^s_{j}, E_j)$
    \STATE $j \gets j + 1$
\UNTIL{$E_j = \texttt{None}$ (no error feedback)}
\STATE $F^s \gets F^s_j$
\STATE
\STATE \textbf{Phase III: Benchmark Evaluation Execution}
\FOR{$d_k \in D$}
    \STATE $\mathcal{P}_{d_k} \gets \mathcal{X}(F^e, M, E_{d_k}^\text{s})$
    \STATE $S_{d_k} \gets \mathcal{X}(F^s, E_{d_k}^\text{s}, \mathcal{P}_{d_k})$
\ENDFOR
\end{algorithmic}
\end{algorithm}

\end{document}